\documentclass[preprint,11pt]{elsarticle}
\usepackage{amsthm}

\usepackage[T1]{fontenc}
\usepackage[utf8]{inputenc}
\usepackage{lmodern}
\usepackage{microtype}
\usepackage{tabularx}
\usepackage{graphicx,booktabs,tabularx,array}

\usepackage{array,ragged2e} 
\newcolumntype{L}{>{\RaggedRight\arraybackslash}X} 
\newcolumntype{C}{>{\centering\arraybackslash}X}   
\newcolumntype{R}{>{\RaggedLeft\arraybackslash}X}  
\usepackage{multicol}
\usepackage{amsmath,amssymb,mathtools,bm}
\usepackage{booktabs,multirow,siunitx}
\usepackage{graphicx,subcaption}
\usepackage{caption}
\usepackage{algorithm}
\usepackage[noend]{algpseudocode}
\usepackage{subcaption}
\usepackage{hyperref}
\usepackage[nameinlink,capitalize]{cleveref}

\biboptions{sort&compress}

\usepackage{amsmath,amssymb,amsthm}
\usepackage{graphicx}
\usepackage{geometry}
\usepackage{parskip}
\usepackage{tikz}
\usepackage{float}
\usepackage{array}
\usepackage{booktabs}
\usepackage{paralist}
\usepackage{pgfplots}
\usepackage{grffile}
\usepackage{hyperref}
\usepackage{subcaption}
\usepackage{natbib}
\pgfplotsset{compat=newest}
\usetikzlibrary{plotmarks}

\usetikzlibrary{positioning,arrows.meta,fit,decorations.pathreplacing}
\usetikzlibrary{calc}
\usepackage[outline]{contour}
\contourlength{1.2pt}
\usepackage{circuitikz}
\usepgfplotslibrary{fillbetween}
\usepackage[x11names, svgnames, rgb]{xcolor}
\usepackage[utf8]{inputenc}
\usepackage{tikz}
\usetikzlibrary{snakes,arrows,shapes}
\usepackage{amsmath}
\usepackage{nicefrac} 
\usepackage{mathtools}
\everymath{\displaystyle} 
\usepackage{rotating}
\journal{arXiv}
\begin{document}
\begin{frontmatter}

\title{Discrete-Time MDP Modeling for Multi-Item Capacitated Lot Sizing with Stochastic Demand Timing}
\author[aff1]{Léa Bayati\corref{cor1}
}
\author[aff2]{Mohamed Dahmoune}
\author[aff1]{Melek Rodoplu}
\cortext[cor1]{Corresponding author. Email: \href{mailto:Lea.Bayati@univ-evry.fr}{Lea.Bayati@univ-evry.fr}}
\address[aff1]{Université Paris-Saclay, Univ Evry, IBISC, 91020 Evry-Courcouronnes, France}
\address[aff2]{Université Paris-Saclay, Univ Evry, 91020 Evry-Courcouronnes, France}

\begin{abstract}
This paper studies a finite-horizon multi-item capacitated lot-sizing problem in which demand quantities are deterministic, while demand-arrival periods are stochastic. Each demand occurs once within a known time window and must be satisfied no later than its deadline. The proposed model makes production and allocation decisions at the demand level, allowing it to represent capacity competition, demand-specific backlog, and allocation-dependent inventory dynamics.
The stochastic problem is formulated as a discrete-time Markov decision process (DTMDP), including the state space, feasible actions, transition kernel, and one-period cost function. To isolate the computational effect of stochastic timing, each stochastic instance is first compared with a deterministic counterpart in which each arrival distribution is replaced by its most likely arrival period. This comparison shows that stochastic timing substantially increases the number of states, the number of transitions, solution time, and memory pressure.
A genetic algorithm (GA) is then proposed for the stochastic-timing problem. The GA searches over feasible state-feedback policies and evaluates each policy exactly under the DTMDP transition model. Computational experiments on 330 benchmark instances show that the GA remains close to the exact stochastic solution whenever the latter is available, with an average optimality gap of about \(3.44\%\). On the difficult benchmark instances, comprising 90 test cases, the GA remains below the \(5\%\) optimality-gap threshold and achieves an average optimization speedup of \(6.89 \pm 1.41\) at the \(95\%\) confidence level. For instances that cannot be solved exactly on the available hardware, an empirical Bellman-time regression is used to estimate the missing exact resolution time and extrapolate the expected GA speedup.
\end{abstract}

\end{frontmatter}
\section{Introduction}
\label{sec:introduction}

Classical lot sizing plans production over a finite horizon by deciding how much to produce in each period and when to set up, under capacity and setup constraints, so as to meet demand at minimum total cost. Most stochastic lot-sizing models focus on uncertainty in demand quantities. However, in several applications, quantities may be contractually fixed while the period in which the demand becomes effective remains uncertain. This situation corresponds to stochastic demand timing. Akartunali et al.~\cite{Akartunali2022EJOR} formalize this uncertainty by assuming that each demand of known quantity occurs in exactly one period of a given interval, according to a discrete probability distribution, and must be satisfied by the end of the interval.

This paper considers a multi-item capacitated setting with setup times, unit production times, setup costs, production costs, holding costs, and backlog costs. An important modeling challenge arises when several demands associated with the same product may arrive simultaneously under limited production capacity. In such situations, aggregate product-level formulations are insufficient because the allocation of available units across demand objects directly affects future backlog costs, deadline feasibility, and inventory evolution. To capture these operational interactions, the proposed framework explicitly models production and allocation decisions at the demand level. This demand-level representation allows the model to distinguish between competing demands sharing the same product and capacity resources, thereby providing a finer dynamic description of stochastic production planning under uncertain demand timing.

The paper has two main methodological contributions. First, we propose a discrete-time Markov decision process formulation for multi-item capacitated lot sizing with stochastic demand timing. Markov decision processes provide a standard framework for sequential decision-making under uncertainty, where decisions are updated as the state of the system evolves over time~\cite{Puterman1994}. The formulation explicitly models the progressive realization of demand arrivals and the resulting adaptive production decisions. Before introducing the heuristic results, the computational study uses deterministic counterparts of the same benchmark instances as a diagnostic baseline. In these counterparts, each stochastic arrival distribution is replaced by a single deterministic arrival period. This deterministic comparison is used to quantify how much difficulty is introduced when moving from deterministic timing to stochastic timing. Second, we introduce a genetic algorithm for stochastic-timing lot sizing that searches over feasible state-feedback policies and evaluates each candidate policy exactly with respect to the DTMDP transition model.

The remainder of the paper is organized as follows. Section~\ref{sec:literature} reviews related work on lot sizing, stochastic demand timing, and genetic algorithms for lot-sizing problems. Section~\ref{sec:problem} defines the problem setting and notation. Section~\ref{sec:mdp} presents the DTMDP formulation. Section~\ref{sec:ga} describes the genetic algorithm heuristic. Section~\ref{sec:complexity} discusses worst-case computational complexity. Section~\ref{sec:experiments} reports the computational experiments, starting with the deterministic-counterpart diagnostic, then the stochastic DTMDP benchmark statistics, performance metrics, optimality-gap analysis, GA speedup analysis, memory observations, and sensitivity analysis. Section~\ref{sec:discussion} discusses the main findings and limitations. Section~\ref{sec:conclusion} concludes the paper.

\section{Literature review}
\label{sec:literature}

The lot-sizing problem is a fundamental topic in operations research, concerned with determining production quantities and timing over a finite planning horizon in order to meet demand at minimum cost. The seminal work of Wagner and Whitin~\cite{wagner1958dynamic} established a dynamic programming framework for the uncapacitated lot-sizing problem and identified key structural properties that enable efficient solutions. Over the years, an extensive body of literature has extended the problem to capacitated and multi-item settings, while also incorporating setup times, setup costs, backlogging, and industrial constraints. For comprehensive surveys and further developments, the interested reader is referred to \cite{brahimi2017single, karimi2003capacitated, quadt2008capacitated}.

Early contributions mainly focused on deterministic settings. To better capture operational uncertainty, a large body of literature has addressed stochastic lot-sizing problems, primarily focusing on uncertainty in demand quantities. Surveys on stochastic lot sizing can be found in \cite{tempelmeier2013stochastic, aloulou2014bibliography}. This stream of literature includes static, dynamic, and static-dynamic uncertainty strategies, where decisions are either fixed in advance or updated over time as uncertainty unfolds. Additional lot-sizing studies have considered other forms of uncertainty and decision coupling, including uncertain operational parameters, random yield, prediction--optimization interactions, and chance-constrained capacitated formulations~\cite{metzker2021optimization,bibak2025integration,gong2023training,deng2026computable}. Despite the richness of this literature, a common limitation is that stochastic demands across periods are often modeled as independent random variables. This can be restrictive when the total demand quantity is known but its timing is uncertain.

To address this limitation, \cite{Akartunali2022EJOR} proposed a model in which demand quantities are deterministic while their occurrence times are uncertain. A given demand amount may occur within a time window spanning several periods, with an associated probability of occurring entirely in each period of that window. As a consequence, demands across periods are correlated: the aggregate quantity over the horizon is fixed, while the timing of its realization is stochastic. Initial contributions focus primarily on single-item or less constrained settings. Extending these ideas to multi-item capacitated systems introduces additional difficulties because demands interact through shared capacity, product-dependent setups, and demand-level allocation decisions \cite{rodoplu2022multi}.

Markov decision processes are a standard framework for sequential decision-making under uncertainty~\cite{Puterman1994}. Recent works by Bayati apply discrete-time MDPs to stochastic operational control problems, including energy-aware data-center management with histogram-based uncertainty and latency considerations~\cite{bayati2018power,bayati2023discrete}. In a similar spirit, this paper uses a finite-horizon DTMDP to model adaptive production and allocation decisions under stochastic demand-arrival timing in multi-item capacitated lot sizing.

Genetic algorithms have also been widely used as metaheuristic solution methods for lot-sizing problems. Guner Goren et al.~\cite{guner2010review} provide a review of GA applications in lot sizing and emphasize that lot-sizing problems are combinatorial and difficult to solve, motivating the use of metaheuristics to obtain good solutions in reasonable computational time. Earlier work by Xie and Dong~\cite{xie2002heuristic} proposed heuristic genetic algorithms for capacitated lot-sizing problems. More recent contributions include hybrid and multi-population genetic algorithms for multi-level capacitated lot-sizing problems with backlogging~\cite{toledo2013hybrid}, genetic-algorithm-based heuristics for capacitated lot-sizing and scheduling problems with sequence-dependent setups~\cite{mohammadi2011genetic,babaei2014genetic}, and adaptive GA variants for multilevel capacitated lot-sizing problems~\cite{wang2022adaptive}. These works support the use of genetic algorithms when exact methods become computationally expensive, particularly in capacitated, multi-item, or multi-level lot-sizing variants.

However, in contrast to these works, the present study focuses on stochastic demand-arrival timing and uses the GA as a policy-search heuristic over the DTMDP state space, with exact DTMDP resolution serving as the benchmark whenever computationally feasible.

Compared with previous expected-cost and anticipative formulations, the present framework explicitly represents the evolution of information over time. Production decisions can adapt to the observed arrival status of each demand, rather than being fixed in advance. Moreover, the demand-level representation captures competition among simultaneous demands for the same product and limited capacity. These features are essential for analyzing capacitated stochastic-timing lot-sizing problems in which allocation decisions affect both feasibility and cost.

\section{Problem description and notation}
\label{sec:problem}
In this section we first summarize the main sets, parameters, and decision variables in Table~\ref{tab:symbols}, and then introduce the problem elements progressively.

\begin{table}[!htb]
\centering
\caption{Main notation (sets, random variables, parameters, decisions, and derived quantities) with $t\in\mathcal{T}$, $k\in\mathcal{K}$, $i\in\mathcal{I}$.}
\label{tab:symbols}
\resizebox{\linewidth}{!}
{%
\begin{tabularx}{1.1\textwidth}{lXl}
\toprule
Symbol & Meaning & Unit \\
\midrule
$\mathcal{T}=\{1,\dots,T\}$ & set of periods (planning horizon) & period index \\
$\mathcal{K}=\{1,\dots,n\}$ & set of products & -- \\
$\mathcal{I}=\{1,\dots,I\}$ & set of demand objects & -- \\
\midrule
$\mathrm{Cap}_t \in \mathbb{R}_+$,\; $t\in\mathcal{T}$
& available capacity in period $t$
& time \\

$\tau^{\mathrm{setup}}_{k,t} \in \mathbb{R}_+$
& setup duration for product $k$ in period $t$
& time \\

$\tau^{\mathrm{unit}}_{k,t} \in \mathbb{R}_+$
& unit processing duration for product $k$ in period $t$
& time/unit \\

$\kappa^{\mathrm{setup}}_{k,t} \in \mathbb{R}_+$
& setup cost for product $k$ in period $t$
& \$ \\

$\kappa^{\mathrm{unit}}_{k,t} \in \mathbb{R}_+$
& unit production cost for product $k$ in period $t$
& \$/unit \\

$h_{k,t} \in \mathbb{R}_+$
& holding cost rate for product $k$ in period $t$
& \$/unit/period \\

$b_{k,t} \in \mathbb{R}_+$
& backlog cost rate for product $k$ in period $t$
& \$/unit/period \\
\midrule

$k_i \in \mathcal{K}$
& product index associated with demand $i$
& product index \\

$Q_i \in \mathbb{N}$
& quantity of demand $i$
& units \\

$s_i,u_i \in \mathcal{T}$, $s_i \le u_i$
& bounds of the timing window of demand $i$ (window $[s_i,u_i]$)
& periods \\

$A_i \in [s_i,u_i]$
& (random) arrival period of demand $i$
& period \\

$p_{i,t} \in [0,1]$
& $\mathbb{P}(A_i=t)$ (defined for $t\in\{s_i,\dots,u_i\}$; $0$ otherwise)
& probability \\
\midrule

$z_{i,t} \in \mathbb{N}$
& decision: units produced in period $t$ allocated to demand $i$
& units \\

$x_{k,t} \in \mathbb{N}$
& derived: total units of product $k$ produced in period $t$
& units \\

$y_{k,t} \in \{0,1\}$
& derived: setup indicator in period $t$ ($y_{k,t}=1$ iff $x_{k,t}>0$)
& -- \\

$P_{i,t} \in \mathbb{N}$
& derived: cumulative allocated units for demand $i$ up to period $t$
& units \\

$\alpha_{i,t} \in \{0,1\}$
& state flag: $1$ if demand $i$ has arrived by the start of period $t$
& -- \\
\bottomrule
\end{tabularx}
}
\end{table}

\paragraph{Horizon and periods}
We use the standard finite-horizon indexing $\mathcal{T}=\{1,2,\dots,T\}$, where each $t\in\mathcal{T}$ is a \emph{period} (or \emph{slot}).

\paragraph{Products and machine capacity}
Let $\mathcal{K}=\{1,\dots,n\}$ be the set of products. Each period $t\in\mathcal{T}$ has a time capacity $\mathrm{Cap}_t$ (time units). If product $k\in\mathcal{K}$ is produced in period $t$ with quantity $x_{k,t}$, then the following time and cost components apply:
\begin{inparaenum}[(i)]
\item \emph{Setup time and cost:} if $x_{k,t}>0$, a setup duration $\tau^{\mathrm{setup}}_{k,t}$ consumes capacity and a fixed setup cost $\kappa^{\mathrm{setup}}_{k,t}$ is incurred;
\item \emph{Processing time and cost:} each produced unit consumes $\tau^{\mathrm{unit}}_{k,t}$ time units and incurs a unit production cost $\kappa^{\mathrm{unit}}_{k,t}$;
\item \emph{Holding cost:} any unit of product $k$ carried in inventory during period $t$ incurs $h_{k,t}$ per unit and per period;
\item \emph{Backlogging cost:} any unit of unmet demand for product $k$ backlogged during period $t$ incurs $b_{k,t}$ per unit and per period.
\end{inparaenum}

\paragraph{Demands as objects with stochastic timing}
We model demand as a set of \emph{demand objects} $\mathcal{I}=\{1,\dots,I\}$.
Each demand $i\in\mathcal{I}$ represents a known order of a specific product and quantity, but with an uncertain realization time: $i$ is a demand for product $k_i$ with deterministic quantity $Q_i$ is assumed to occur (arrive) at a random period $A_i$ within its feasible window $[s_i,u_i]$, according to the discrete probabilities defined below.
Each demand $i\in\mathcal{I}$ is defined by:
\begin{inparaenum}[(i)]
\item a product index $k_i\in\mathcal{K}$;
\item a deterministic quantity $Q_i\in\mathbb{N}$ (assumed known);
\item a time window $[s_i,u_i]\subseteq\mathcal{T}$;
\item a discrete timing distribution over the window: $p_{i,t}=\mathbb{P}(A_i=t), \qquad t\in\{s_i,\dots,u_i\}, \qquad \sum_{t=s_i}^{u_i} p_{i,t}=1,$ where $A_i$ is the (random) arrival period of demand $i$.
\end{inparaenum}

\paragraph{Service requirement (deadline)}
Demand $i$ must be fully satisfied no later than period $u_i$.
Equivalently, production allocated to demand $i$ is not allowed after $u_i$.

\paragraph{Deterministic timing as a Dirac histogram}
If demand $i$ is known to occur at a fixed period $t_i$, then its window is the singleton
$[s_i,u_i]=[t_i,t_i]$
and its distribution is the Dirac histogram:
$
p_{i,t_i}=1,\qquad p_{i,t}=0 \ \text{for}\ t\neq t_i .
$
This unifies deterministic and stochastic timing in one notation.

\paragraph{Why per-demand allocation is needed (simultaneous arrivals)}
When multiple demands for the same product occur in the same period, the available units (inventory plus current production) may be insufficient to satisfy all of them.
The way we allocate units across these demands changes which demand is backlogged and for how long, which changes the total cost.

\paragraph{Example (two demands for the same product)}
Consider two demands $i=1,2$ with the same product $k_1=k_2=k$.
Let $Q_1=30$ with window $[s_1,u_1]=[3,7]$ and $Q_2=20$ with window $[s_2,u_2]=[1,5]$.
In a scenario where both arrivals occur at $t=4$, suppose only 5 units of product $k$ are available at that time.
A decision is needed to allocate these 5 units between demands 1 and 2.
This motivates a per-demand decision variable.

\paragraph{Data parameters vs.\ company parameters}
For each demand $i\in\mathcal{I}$, parameters $(k_i,Q_i,s_i,u_i,p_{i,t})$ are treated as input demand-timing information. However, machine parameters $(\mathrm{Cap}_t)_{t\in\mathcal{T}}$, $(\tau^{\mathrm{setup}}_{k,t},\tau^{\mathrm{unit}}_{k,t})_{k\in\mathcal{K},t\in\mathcal{T}}$, $(\kappa^{\mathrm{setup}}_{k,t},\kappa^{\mathrm{unit}}_{k,t})_{k\in\mathcal{K},t\in\mathcal{T}}$, and cost rates $(h_{k,t},b_{k,t})_{k\in\mathcal{K},t\in\mathcal{T}}$ are company-specific.
In practice they can be defined from:
\begin{inparaenum}[(i)]
\item routing sheets / standard times for $\tau^{\mathrm{unit}}_{k,t}$;
\item changeover logs for $\tau^{\mathrm{setup}}_{k,t}$;
\item accounting/controlling for $\kappa^{\mathrm{setup}}_{k,t}$ and $\kappa^{\mathrm{unit}}_{k,t}$;
\item inventory carrying rate (space, capital, spoilage) for $h_{k,t}$;
\item service-level targets or contract penalties for $b_{k,t}$.
\end{inparaenum}

\section{Discrete-time MDP formulation}
\label{sec:mdp}

We now define a DTMDP that captures the nature of decisions under stochastic demand arrivals.
The formulation follows the standard $(\mathcal{S},\mathcal{A},\mathbb{P},r)$ structure of an MDP.

\begin{table}[!htb]
\centering
\caption{DTMDP components for the lot-sizing problem.}
\label{tab:mdp-components}
\resizebox{\linewidth}{!}{%
\begin{tabularx}{1.3\textwidth}{lL}
\toprule
\bf{Component} & \bf{Description} \\
\midrule
Decision epochs & Periods $t\in\mathcal{T}=\{1,\dots,T\}$ (finite horizon). \\\midrule
State $\mathcal{S}$& Composed of period index $t$, arrival flags $(\alpha_{i,t})_{i\in\mathcal{I}}$, and cumulative allocations $(P_{i,t-1})_{i\in\mathcal{I}}$ (see \eqref{eq:state}). \\\midrule
Action $\mathcal{A}$& Composed of  per-demand production/allocation vector $a_t=(z_{i,t})_{i\in\mathcal{I}}$ subject to capacity and feasibility constraints (see \eqref{eq:action}--\eqref{eq:cap}). \\\midrule
Transition dynamics $\mathbb{P}$& Deterministic update of $P_{i,t}$ given the action, followed by stochastic arrival updates driven by hazards $q_{i,t}$ (see \eqref{eq:Pupdate}--\eqref{eq:arrupdate} and \eqref{eq:trans-kernel}). \\\midrule
One-step cost $r$& Setup/production costs plus holding/backlog costs incurred during period $t$ (see \eqref{eq:stage}). \\\midrule
Objective & Minimize the expected total cost over the horizon subject to deadlines (see \eqref{eq:deadline-eq}) and Bellman recursion (see \eqref{eq:bellman}). \\
\bottomrule
\end{tabularx}
}
\end{table}

\subsection{Arrival process: step-by-step hazards}
\label{sec:hazards}

The timing information available from data is a \emph{marginal} arrival pmf $(p_{i,t})_{t\in\mathcal{T}}$, where $p_{i,t}=\mathbb{P}(A_i=t)$ and $p_{i,t}=0$ outside the window $[s_i,u_i]$.
When we model time explicitly and let it advance one period at a time, the one-step transition probabilities cannot use $p_{i,t}$ directly because, at the beginning of period $t$, we already know whether the demand has arrived before $t$.
To obtain a period-by-period DTMC consistent with the same marginal pmf $(p_{i,t})$, we therefore use the associated \emph{conditional} arrival probabilities (hazards).
For each demand $i\in\mathcal{I}$ and each $t\in\{s_i,\dots,u_i\}$, define
\begin{equation}
q_{i,t}
\;=\;
\mathbb{P}(A_i=t \mid A_i \ge t)
\;=\;
\frac{p_{i,t}}{1-\displaystyle\sum_{\tau=s_i}^{t-1} p_{i,\tau}},
\label{eq:hazard}
\end{equation}
with the convention that $q_{i,u_i}=1$ when $\sum_{t=s_i}^{u_i}p_{i,t}=1$ (i.e., the demand must arrive no later than $u_i$).
By construction, these hazards reproduce the original marginal pmf when time is unfolded step by step.

\paragraph{DTMC state encoding}
We represent the arrival evolution with DTMC states of the form $s=(t,\alpha)$, where $t$ is the current period index and $\alpha\in\{0,1\}$ indicates whether the demand has arrived by the start of period $t$ ($\alpha=1$ means ``already arrived'').
From a non-arrived state $(t,0)$ with $t\le u_i$, the demand arrives during period $t$ with probability $q_{i,t}$, leading to $(t+1,1)$, and does not arrive with probability $1-q_{i,t}$, leading to $(t+1,0)$ (except at $t=u_i$ where arrival is forced).
From an arrived state $(t,1)$, the process deterministically moves to $(t+1,1)$.

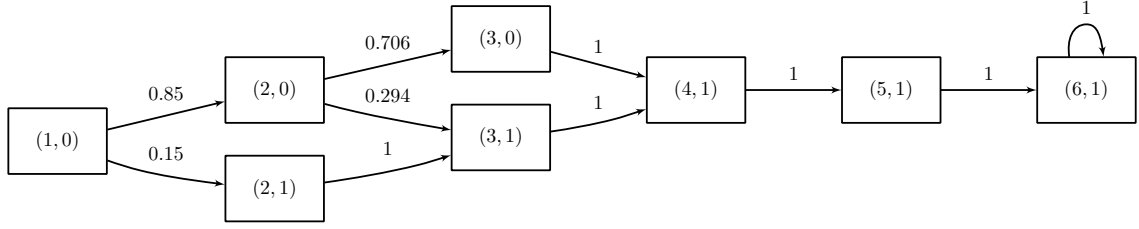
\begin{figure}[!htb]
\centering
\resizebox{1\linewidth}{!}{
\begin{tikzpicture}[>=latex',line join=bevel,]
  \pgfsetlinewidth{1bp}
\pgfsetcolor{black}
  \draw [->] (54.257bp,50.284bp) .. controls (70.365bp,54.139bp) and (91.215bp,59.128bp)  .. (118.83bp,65.737bp);
  \definecolor{strokecol}{rgb}{0.0,0.0,0.0};
  \pgfsetstrokecolor{strokecol}
  \draw (86.5bp,70.0bp) node {$0.85$};
  \draw (60.257bp,44.284bp) node {$$};
  \draw (112.83bp,59.737bp) node {$$};
  \draw [->] (54.242bp,33.428bp) .. controls (60.021bp,31.39bp) and (66.161bp,29.444bp)  .. (72.0bp,28.0bp) .. controls (83.813bp,25.078bp) and (96.928bp,22.947bp)  .. (118.73bp,20.234bp);
  \draw (86.5bp,37.0bp) node {$0.15$};
  \draw (60.242bp,27.428bp) node {$$};
  \draw (112.73bp,14.234bp) node {$$};
  \draw [->] (173.17bp,78.007bp) .. controls (190.62bp,82.011bp) and (213.79bp,87.328bp)  .. (242.87bp,94.003bp);
  \draw (208.0bp,98.0bp) node {$0.706$};
  \draw (179.17bp,72.007bp) node {$$};
  \draw (236.87bp,88.003bp) node {$$};
  \draw [->] (173.43bp,64.384bp) .. controls (179.2bp,62.836bp) and (185.28bp,61.29bp)  .. (191.0bp,60.0bp) .. controls (204.55bp,56.941bp) and (219.56bp,54.117bp)  .. (242.88bp,50.093bp);
  \draw (208.0bp,69.0bp) node {$0.294$};
  \draw (179.43bp,58.384bp) node {$$};
  \draw (236.88bp,44.093bp) node {$$};
  \draw [->] (173.22bp,21.28bp) .. controls (188.46bp,23.456bp) and (207.96bp,26.706bp)  .. (225.0bp,31.0bp) .. controls (227.62bp,31.661bp) and (230.32bp,32.407bp)  .. (242.69bp,36.248bp);
  \draw (208.0bp,40.0bp) node {$1$};
  \draw (179.22bp,15.28bp) node {$$};
  \draw (236.69bp,30.248bp) node {$$};
  \draw [->] (297.23bp,93.007bp) .. controls (310.13bp,89.566bp) and (325.89bp,85.364bp)  .. (349.75bp,79.001bp);
  \draw (323.5bp,96.0bp) node {$1$};
  \draw (303.23bp,87.007bp) node {$$};
  \draw (343.75bp,73.001bp) node {$$};
  \draw [->] (297.05bp,49.179bp) .. controls (307.94bp,50.79bp) and (320.68bp,53.042bp)  .. (332.0bp,56.0bp) .. controls (334.64bp,56.689bp) and (337.34bp,57.472bp)  .. (349.73bp,61.544bp);
  \draw (323.5bp,65.0bp) node {$1$};
  \draw (303.05bp,43.179bp) node {$$};
  \draw (343.73bp,55.544bp) node {$$};
  \draw [->] (404.23bp,72.0bp) .. controls (417.0bp,72.0bp) and (432.58bp,72.0bp)  .. (456.75bp,72.0bp);
  \draw (430.5bp,81.0bp) node {$1$};
  \draw (410.23bp,66.0bp) node {$$};
  \draw (450.75bp,66.0bp) node {$$};
  \draw [->] (511.23bp,72.0bp) .. controls (524.0bp,72.0bp) and (539.58bp,72.0bp)  .. (563.75bp,72.0bp);
  \draw (537.5bp,81.0bp) node {$1$};
  \draw (517.23bp,66.0bp) node {$$};
  \draw (557.75bp,66.0bp) node {$$};
  \draw [->] (581.59bp,90.153bp) .. controls (580.15bp,99.539bp) and (583.28bp,108.0bp)  .. (591.0bp,108.0bp) .. controls (595.7bp,108.0bp) and (598.71bp,104.86bp)  .. (600.41bp,90.153bp);
  \draw (591.0bp,117.0bp) node {$1$};
  \draw (575.59bp,96.153bp) node {$$};
  \draw (594.41bp,96.153bp) node {$$};
\begin{scope}
  \definecolor{strokecol}{rgb}{0.0,0.0,0.0};
  \pgfsetstrokecolor{strokecol}
  \draw (54.0bp,62.0bp) -- (0.0bp,62.0bp) -- (0.0bp,26.0bp) -- (54.0bp,26.0bp) -- cycle;
  \draw (27.0bp,44.0bp) node {$(1,0)$};
\end{scope}
\begin{scope}
  \definecolor{strokecol}{rgb}{0.0,0.0,0.0};
  \pgfsetstrokecolor{strokecol}
  \draw (173.0bp,90.0bp) -- (119.0bp,90.0bp) -- (119.0bp,54.0bp) -- (173.0bp,54.0bp) -- cycle;
  \draw (146.0bp,72.0bp) node {$(2,0)$};
\end{scope}
\begin{scope}
  \definecolor{strokecol}{rgb}{0.0,0.0,0.0};
  \pgfsetstrokecolor{strokecol}
  \draw (173.0bp,36.0bp) -- (119.0bp,36.0bp) -- (119.0bp,0.0bp) -- (173.0bp,0.0bp) -- cycle;
  \draw (146.0bp,18.0bp) node {$(2,1)$};
\end{scope}
\begin{scope}
  \definecolor{strokecol}{rgb}{0.0,0.0,0.0};
  \pgfsetstrokecolor{strokecol}
  \draw (297.0bp,118.0bp) -- (243.0bp,118.0bp) -- (243.0bp,82.0bp) -- (297.0bp,82.0bp) -- cycle;
  \draw (270.0bp,100.0bp) node {$(3,0)$};
\end{scope}
\begin{scope}
  \definecolor{strokecol}{rgb}{0.0,0.0,0.0};
  \pgfsetstrokecolor{strokecol}
  \draw (297.0bp,64.0bp) -- (243.0bp,64.0bp) -- (243.0bp,28.0bp) -- (297.0bp,28.0bp) -- cycle;
  \draw (270.0bp,46.0bp) node {$(3,1)$};
\end{scope}
\begin{scope}
  \definecolor{strokecol}{rgb}{0.0,0.0,0.0};
  \pgfsetstrokecolor{strokecol}
  \draw (404.0bp,90.0bp) -- (350.0bp,90.0bp) -- (350.0bp,54.0bp) -- (404.0bp,54.0bp) -- cycle;
  \draw (377.0bp,72.0bp) node {$(4,1)$};
\end{scope}
\begin{scope}
  \definecolor{strokecol}{rgb}{0.0,0.0,0.0};
  \pgfsetstrokecolor{strokecol}
  \draw (511.0bp,90.0bp) -- (457.0bp,90.0bp) -- (457.0bp,54.0bp) -- (511.0bp,54.0bp) -- cycle;
  \draw (484.0bp,72.0bp) node {$(5,1)$};
\end{scope}
\begin{scope}
  \definecolor{strokecol}{rgb}{0.0,0.0,0.0};
  \pgfsetstrokecolor{strokecol}
  \draw (618.0bp,90.0bp) -- (564.0bp,90.0bp) -- (564.0bp,54.0bp) -- (618.0bp,54.0bp) -- cycle;
  \draw (591.0bp,72.0bp) node {$(6,1)$};
\end{scope}
\end{tikzpicture}
}
\caption{DTMC for Arrival1 using step-by-step hazards. Node label \texttt{s\,(t,$\alpha$)}: \texttt{s} is the DTMC state id, $t$ is the current period index (starting at $1$), and $\alpha\in\{0,1\}$ indicates whether the demand has arrived by the start of period $t$. Transition probabilities are built from the conditional hazards in~\eqref{eq:hazard}, yielding the marginal pmf in~\eqref{eq:arrival1-pmf}.}
\label{fig:arrival1-dtmc-hazard}
\end{figure}

\paragraph{Example}
Assume periods start at $1$ until $6$ and consider a demand with window $[1,3]$ and marginal pmf
\begin{equation}
p_{1}=0.15,\qquad p_{2}=0.25,\qquad p_{3}=0.60.
\label{eq:arrival1-pmf}
\end{equation}
Using~\eqref{eq:hazard}, the corresponding hazards are
\begin{equation*}
q_{1}=0.15,\qquad
q_{2}=\frac{0.25}{1-0.15}=\frac{0.25}{0.85}\approx 0.294,\qquad
q_{3}=1.
\end{equation*}
These hazards preserve the marginal distribution:
\begin{equation*}
\left\{
\begin{aligned}
\mathbb{P}(A=1) &= q_1 = 0.15,\\
\mathbb{P}(A=2) &= (1-q_1)q_2 = 0.85 \times 0.294 \approx 0.25,\\
\mathbb{P}(A=3) &= (1-q_1)(1-q_2)\times 1 = 0.60.
\end{aligned}
\right.
\end{equation*}
Figure~\ref{fig:arrival1-dtmc-hazard} shows the resulting DTMC.
Each node is labeled by the DTMC state id and the pair $(t,\alpha)$, where $t$ is the period index (starting at $1$) and $\alpha=1$ indicates that the demand has already arrived.

\subsection{State space $\mathcal{S}$}
At the beginning of period $t$, the information available to the decision maker is:
\begin{inparaenum}[(i)]
\item the current period $t$;
\item for each demand $i$, an arrival flag $\alpha_{i,t}\in\{0,1\}$ indicating whether demand $i$ has already arrived by the start of period $t$;
\item for each demand $i$, the cumulative allocated production $P_{i,t-1}\in\{0,1,\dots,Q_i\}$ produced for demand $i$ up to the end of period $t-1$.
\end{inparaenum}
Thus a state can be written as
\begin{equation}
s_t = \big(t,\ (\alpha_{i,t})_{i\in\mathcal{I}},\ (P_{i,t-1})_{i\in\mathcal{I}}\big).
\label{eq:state}
\end{equation}

The initial state at the beginning of the horizon is
$
s_1=\big(1,\ (\alpha_{i,1}=0)_{i\in\mathcal{I}},\ (P_{i,0}=0)_{i\in\mathcal{I}}\big),
$
i.e., no demand has arrived yet and no production has been allocated.

Notice that we do not need an explicit inventory state, we restrict decisions to produce only for existing demands (no extra production beyond $Q_i$ for demande $i$), and we track production per demand.
Then, units produced for a demand that has not arrived yet represent inventory for that demand.
Hence the inventory of product $k$ at the start of period $t$ is derivable as
$
\mathrm{Inv}_{k,t}=\sum_{i\in\mathcal{I}:k_i=k} (1-\alpha_{i,t})\,P_{i,t-1},
$
and does not need an additional state component.

Here, inventory is interpreted as demand-dedicated pre-production. Since production decisions are tracked at the demand level and each demand occurs only once, units produced for a demand that has not yet arrived correspond to reserved inventory associated with that demand object. Consequently, cumulative allocated production and inventory coincide in the present formulation.

\subsection{Action space $\mathcal{A}$}
In period $t$, the decision is how many units to produce for each demand:
\begin{equation}
a_t = (z_{i,t})_{i\in\mathcal{I}},
\qquad
z_{i,t}\in\{0,1,\dots,Q_i-P_{i,t-1}\}.
\label{eq:action}
\end{equation}
The induced per-product production is
\begin{equation}
x_{k,t}=\sum_{i\in\mathcal{I}:k_i=k} z_{i,t},
\qquad
y_{k,t}=\begin{cases}
1, & x_{k,t}>0,\\
0, & x_{k,t}=0.
\end{cases}
\end{equation}
%
An action $a_t$ is feasible if it satisfies:
\begin{equation}
\sum_{k\in\mathcal{K}}\Big(\tau^{\mathrm{setup}}_{k,t}\,y_{k,t}+\tau^{\mathrm{unit}}_{k,t}\,x_{k,t}\Big)
\le \mathrm{Cap}_t.
\label{eq:cap}
\end{equation}

\subsection{Transition kernel $\mathbb{P}$}
Given a state $s_t$ and an action $a_t$, the next state $s_{t+1}$ is generated in two steps.

\paragraph{(1) Deterministic production update}
First, cumulative allocations update deterministically:
\begin{equation}
P_{i,t}=P_{i,t-1}+z_{i,t}, \qquad \forall i\in\mathcal{I}.
\label{eq:Pupdate}
\end{equation}

\paragraph{(2) Stochastic arrival update}
Then, for each demand $i$ such that $\alpha_{i,t}=0$, if $t\in\{s_i,\dots,u_i\}$:
\begin{equation}
\mathbb{P}(\alpha_{i,t+1}=1 \mid \alpha_{i,t}=0) = q_{i,t},
\qquad
\mathbb{P}(\alpha_{i,t+1}=0 \mid \alpha_{i,t}=0)=1-q_{i,t}.
\label{eq:arrupdate}
\end{equation}
Once a demand has arrived, it remains arrived, i.e., $\alpha_{i,t}=1$ implies $\alpha_{i,t+1}=1$.

We assume demands are independent; hence, conditional on $(s_t,a_t)$, the joint transition probability factorizes across demands. More precisely, let $s_t$ be given by~\eqref{eq:state} and let $s_{t+1}=(t+1,(\alpha_{i,t+1})_{i\in\mathcal{I}},(P_{i,t})_{i\in\mathcal{I}})$ where $P_{i,t}=P_{i,t-1}+z_{i,t}\ \ \forall i\in\mathcal{I}$. Then
\begin{equation}
\mathbb{P}(s_{t+1}\mid s_t,a_t)
=
\prod_{i\in\mathcal{I}} \pi_{i,t}(\alpha_{i,t+1}\mid \alpha_{i,t}).
\label{eq:trans-kernel}
\end{equation}
where, for each $i\in\mathcal{I}$,
\begin{equation}
\pi_{i,t}(\alpha'\mid \alpha)=
\begin{cases}
1, & \alpha=1\ \text{and}\ \alpha'=1,\\
q_{i,t}, & \alpha=0,\ \alpha'=1,\ t\in\{s_i,\dots,u_i\},\\
1-q_{i,t}, & \alpha=0,\ \alpha'=0,\ t\in\{s_i,\dots,u_i\},\\
1, & \alpha=0,\ \alpha'=0,\ t\notin\{s_i,\dots,u_i\}.
\end{cases}
\end{equation}

\subsection{Stage cost $r$}
We define the stage cost for period $t$ as the sum of:
\begin{inparaenum}[(i)]
\item setup and production costs induced by $(x_{k,t},y_{k,t})$;
\item holding cost for units produced for demands that have not arrived by the end of the period;
\item backlog cost for unmet quantities of demands that have arrived by the end of the period.
\end{inparaenum}

Using the post-action cumulative $P_{i,t}$ and the next arrival flags $\alpha_{i,t}$, define the holding and backlog units for each demand:
\begin{equation}
H_{i,t}=
\begin{cases}
P_{i,t}, & \alpha_{i,t}=0,\\
0, & \alpha_{i,t}=1,
\end{cases}
\qquad
B_{i,t}=
\begin{cases}
0, & \alpha_{i,t}=0,\\
\max(0,Q_i-P_{i,t}), & \alpha_{i,t}=1.
\end{cases}
\end{equation}
The stage cost is then:
\begin{equation}
r(s_t,a_t)=
\sum_{k\in\mathcal{K}}\Big(\kappa^{\mathrm{setup}}_{k,t}\,y_{k,t}+\kappa^{\mathrm{unit}}_{k,t}\,x_{k,t}\Big)
+\sum_{i\in\mathcal{I}}\Big(h_{k_i,t}\,H_{i,t}+b_{k_i,t}\,B_{i,t}\Big).
\label{eq:stage}
\end{equation}

\subsection{Deadlines and feasibility}
The deadline requirement ``satisfy by $u_i$'' can be enforced as a hard constraint by forbidding production after $u_i$ and ensuring completion at $u_i$:
\begin{equation}
P_{i,u_i}=Q_i, \qquad \forall i\in\mathcal{I}.
\label{eq:deadline-eq}
\end{equation}
In practice, the DTMDP can implement this by restricting actions $z_{i,t}=0$ for all $t>u_i$.

\begin{sidewaysfigure}[p]
\centering
\includegraphics[width=1.00\linewidth]{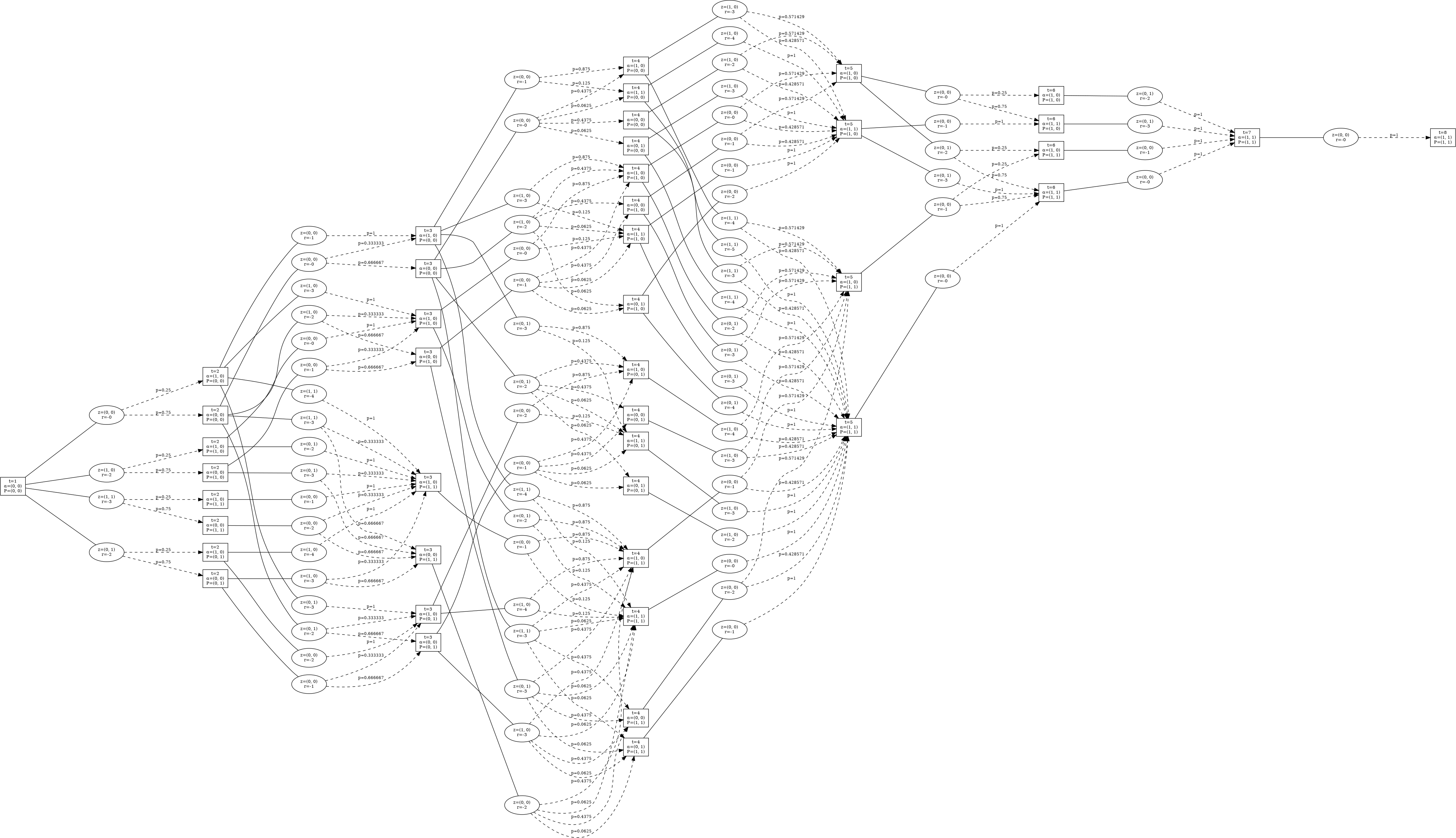}
\caption{Illustrative DTMDP transition graph for a two-product instance. The instance contains two unit demands associated with two distinct products. Demand \(D_1\), associated with product \(P_1\), has arrival window \([1,4]\) and probability distribution \((0.25,0.25,0.25,0.25,0,0,0)\). Demand \(D_2\), associated with product \(P_2\), has arrival window \([3,6]\) and probability distribution \((0,0,0.125,0.375,0.375,0.125,0)\). Rectangular nodes represent decision states with feasible production actions, oval nodes represent stochastic arrival transitions, and dashed arcs indicate transition probabilities between successive states.}
\label{fig:dtmdp-two-product-example}
\end{sidewaysfigure}

Figure~\ref{fig:dtmdp-two-product-example} illustrates the resulting DTMDP structure on a small two-product instance with two stochastic demands, showing how decision states, feasible production actions, and probabilistic arrival transitions are combined over the planning horizon.


\subsection{Bellman optimality and solution}
Let $V_t(s)$ be the optimal cost-to-go from state $s$ at the start of period $t$.
The finite-horizon Bellman recursion is:
\begin{equation}
V_t(s)=\min_{a\in\mathcal{A}(s)}\left\{r(s,a)+\sum_{s'} \mathbb{P}(s' \mid s,a)\,V_{t+1}(s')\right\},
\qquad V_{T+1}(s)=0.
\label{eq:bellman}
\end{equation}
This recursion can be solved by backward induction.
For larger instances or variants, value iteration and policy iteration can be considered \cite{Puterman1994}.


\section{Genetic algorithm heuristic}
\label{sec:ga}

The exact DTMDP formulation provides an optimal policy, but its explicit state--action representation may become large when the number of demands, the demand quantities, or the planning horizon increase. We therefore implement a genetic algorithm (GA) heuristic that searches directly over feasible state-feedback policies. The heuristic uses the same state space, action set, transition matrices, and one-period rewards as the exact DTMDP. Hence, the comparison between the exact method and the GA is performed on the same model and the same objective function.

\subsection{Solution representation}

An individual represents a complete policy over the reachable states of the DTMDP. Let \(\mathcal{S}\) be the set of states generated during the MDP construction, and let \(\mathcal{A}(s)\) be the set of feasible actions in state \(s\). A chromosome is defined as
$
\chi=(\chi_s)_{s\in\mathcal{S}},
$
where each gene \(\chi_s\in\mathcal{A}(s)\) is the action selected when state \(s\) is visited. Since the period index is part of the state, this state-based encoding is sufficient to represent a non-stationary finite-horizon policy. In the implementation, a gene stores the index of a feasible action in the global action catalogue. Feasibility is enforced by precomputing, for every state, the list of admissible action indices satisfying production bounds, capacity constraints, and deadline constraints.

The action associated with a state is a vector
$
a_t=(z_{i,t})_{i\in\mathcal{I}},
$
where \(z_{i,t}\) is the quantity produced at period \(t\) and allocated to demand \(i\). The induced product-level quantities \(x_{k,t}\) and setup indicators \(y_{k,t}\) are derived exactly as in the MDP formulation.

\subsection{Fitness evaluation}

The fitness of a chromosome is the expected total reward of the policy that it encodes. Since the model is a cost minimization problem and the implementation uses rewards equal to negative costs, maximizing fitness is equivalent to minimizing expected cost. Starting from the initial state \(s_1\), the probability mass over states is propagated over the finite horizon using the MDP transition kernel. For a chromosome \(\chi\), the fitness is
\begin{equation}
F(\chi)
=
\mathbb{E}_{\chi}\left[\sum_{t=1}^{T} R(s_t,\chi_{s_t})\right]
=
-\mathbb{E}_{\chi}\left[\sum_{t=1}^{T} c(s_t,\chi_{s_t})\right],
\end{equation}
where \(R(s,a)=-c(s,a)\) denotes the one-period reward associated with state-action pair \((s,a)\). The transition probabilities are not resampled during evaluation; instead, the expected value is computed exactly by propagating the state distribution. This makes the fitness deterministic for a fixed chromosome.

\subsection{Population initialization}
\label{sec:ga-initialization}

The initial population combines a small number of structured policies with randomly generated feasible policies. The structured part contains simple constructive rules, such as a myopic policy based on immediate reward, a maximum-production policy, a minimum-production policy, and an urgency-based policy. Random individuals are generated by sampling one feasible action independently for each state. This hybrid initialization provides both feasible diversity and a small set of informative starting policies.

The urgency-based seed uses the temporal information of the stochastic demands. For a state $s_t=(t,\alpha_t,P_{t-1})$ and an action $a=(z_i)_{i\in\mathcal{I}}$, the score is
\begin{equation}
\mathrm{score}(s_t,a)
=
\sum_{i\in\mathcal{I}:z_i>0}
\frac{
z_i\left(1+\alpha_{i,t}\right)\left(1+p_{i,t}\right)
}{
\max(1,Q_i-P_{i,t-1})\max(1,u_i-t+1)
}.
\end{equation}
This score favours actions that produce a large fraction of the remaining quantity, especially for demands that have already arrived, are close to their deadline, or have a high probability of arriving in the current period. The score is used only for initialization; it is not part of the DTMDP objective function and does not affect policy evaluation.

A limited fraction of the population may also be initialized from relaxed DTMDP policies. These auxiliary policies are obtained by solving simplified variants of the original DTMDP, for example with coarser time aggregation, coarser production discretization, a smaller action catalogue, or demand aggregation by product. The relaxed policies are projected back onto the original state space and inserted only as initial chromosomes. All subsequent evaluation and selection are performed on the original DTMDP.

\subsection{Parent pairing and offspring generation}

At each generation, the population of size $N$ is randomly shuffled and divided into $N/2$ parent pairs. Each pair generates exactly four children. Two children are produced by gene-uniform crossover and two children are produced by time-block crossover. This deterministic offspring count is used in all experiments and removes the need for an additional offspring-size parameter.

\paragraph{Gene-uniform crossover}
For each state gene independently, the child inherits the action of the first parent with probability $0.5$ and the action of the second parent with probability $0.5$:
\begin{equation}
\chi^{\mathrm{child}}_s=
\begin{cases}
\chi^{\mathrm{father}}_s, & \text{with probability }0.5,\\
\chi^{\mathrm{mother}}_s, & \text{with probability }0.5.
\end{cases}
\end{equation}
This operator is exploratory because it recombines parent policies at the state level.

\paragraph{Time-block crossover}
A cut period $\bar t\in\{1,\dots,T\}$ is sampled uniformly. The child inherits all genes associated with states whose period is at most $\bar t$ from one parent, and the remaining genes from the other parent. This operator preserves part of the temporal structure of a policy.

\subsection{Mutation}

Mutation is applied at the child level. With probability $\mu$, a child is mutated. When mutation is triggered, a fraction $\phi$ of its genes is selected uniformly at random, and each selected gene is replaced by a randomly sampled feasible action for the corresponding state. Thus, mutation never creates infeasible genes. Formally, for a selected state $s$,
\begin{equation}
\chi_s \leftarrow \mathrm{Uniform}(\mathcal{A}(s)).
\end{equation}
The parameters $\mu$ and $\phi$ control, respectively, how often mutation is applied and how strongly a mutated child is modified.

\subsection{Selection and stopping rule}

The offspring generated at a generation are evaluated using the exact DTMDP transition model. The next population is then built from the evaluated offspring by retaining a fraction of the best children and completing the population with randomly selected remaining children. The best chromosome found over all generations is stored separately as the incumbent solution. This selection rule preserves high-quality offspring while maintaining some diversity between generations.

The algorithm stops after a fixed maximum number of generations $G$, or earlier if convergence is detected. Convergence is declared when the best global fitness has not improved by more than a tolerance $\varepsilon$ for a prescribed number $G_{\mathrm{stall}}$ of consecutive generations.

\begin{algorithm}[!htb]
\caption{Genetic algorithm for the stochastic-timing lot-sizing DTMDP}
\label{alg:ga-lotsizing}
\begin{algorithmic}[1]
\Require State set $\mathcal{S}$
\Require feasible action lists $\mathcal{A}(s)$
\Require transition matrices $P$
\Require reward matrix $R$
\Require population size $N$
\Require maximum generations $G$
\Require mutation parameters $(\mu,\phi)$
\Ensure Best chromosome $\chi^\star$ and its expected cost
\State Generate an initial population $\mathcal{P}_0$ of size $N$
\State Evaluate every chromosome in $\mathcal{P}_0$ using the DTMDP transition model
\State Set $\chi^\star$ to the best chromosome found so far
\For{generation $g=1,\dots,G$}
    \State Randomly shuffle $\mathcal{P}_{g-1}$ and form $N/2$ parent pairs
    \State Initialize offspring set $\mathcal{C}_g\gets\emptyset$
    \For{each parent pair $(\chi^{p},\chi^{m})$}
        \State Generate two children by gene-uniform crossover
        \State Generate two children by time-block crossover
        \State Mutate each generated child with probability $\mu$ and,
        \State if mutation occurs, replace a fraction $\phi$ of its genes
        \State Add the four children to $\mathcal{C}_g$
    \EndFor
    \State Evaluate all children in $\mathcal{C}_g$
    \State Update $\chi^\star$ if a better chromosome is found
    \State Build $\mathcal{P}_g$ from the best children and randomly selected remaining children
    \If{no improvement greater than $\varepsilon$ is observed for $G_{\mathrm{stall}}$ generations} \textbf{ break}
    \EndIf
\EndFor
\State \Return $\chi^\star$
\end{algorithmic}
\end{algorithm}

\section{Computational complexity}
\label{sec:complexity}
This section gives simple worst-case complexity bounds. The objective is not to provide tight bounds, but to identify the dominant growth factors of the exact DTMDP and of the genetic algorithm. The bounds are expressed directly in terms of the planning horizon $T$, the number of demands $I$, the number of products $K$, the largest demand quantity $Q_{\max}$, the GA population size $N$, and the number of generations $G$.
Let $Q_{\max}=\max_{i=1,\ldots,I}Q_i$. Each demand has one binary arrival-status variable and one cumulative production variable. Hence, before reachability pruning, a conservative upper bound on the number of DTMDP states is $\mathcal{O}\left(T\,2^I\,Q_{\max}^I\right)$. The factor $2^I$ corresponds to the possible arrival-status vectors, while $Q_{\max}^I$ bounds the cumulative production vectors. The notation uses $Q_{\max}^I$ instead of $(Q_{\max}+1)^I$, since the two expressions are equivalent in Big-O notation for $Q_{\max}\ge 1$.
In the worst case, an action specifies one production quantity for each demand. Therefore, a crude upper bound on the action catalogue size is $\mathcal{O}\left(Q_{\max}^I\right)$. In practice, this catalogue is reduced by capacity constraints, deadline constraints, production discretization, and explicit catalogue-size limits.

\subsection{Exact DTMDP}
The DTMDP construction scans actions for reachable states, checks demand-level and product-level feasibility, and enumerates the possible stochastic arrival outcomes. The number of arrival outcomes in one period is bounded by $2^I$. Thus, a conservative construction-time bound is $\mathcal{O}\left(T\,2^I\,Q_{\max}^{2I}\,(I+n+2^I)\right)$. The memory required to store the DTMDP is dominated by the sparse transition structure. Since each state-action pair may have up to $2^I$ stochastic successors, the transition-storage bound is $\mathcal{O}\left(T\,4^I\,Q_{\max}^{2I}\right)$.
The finite-horizon exact solver used in the experiments applies a backward dynamic-programming recursion on the constructed DTMDP. With the generic finite-horizon representation, the time complexity is bounded by $\mathcal{O}\left(T^2\,4^I\,Q_{\max}^{2I}\right)$. The additional memory of the value and policy tables is bounded by $\mathcal{O}\left(T^2\,2^I\,Q_{\max}^{I}\right)$, although in large instances the transition model itself is usually the dominant memory component.

%
%

\subsection{Genetic algorithm}
A chromosome fixes one feasible action per reachable state. Therefore, evaluating a chromosome does not scan the full action catalogue. It propagates the probability distribution over the DTMDP under the fixed policy. Since the state set already includes the time index, the cost of evaluating one chromosome is bounded by the number of reachable states times the stochastic branching factor, namely $\mathcal{O}\left(T\,4^I\,Q_{\max}^{I}\right)$.
Each generation produces exactly four children per parent pair. Since there are $N/2$ parent pairs, the number of children evaluated per generation is $2N$. Up to constant factors, the full GA run over $G$ generations therefore has time complexity $\mathcal{O}\left(G\,N\,T\,4^I\,Q_{\max}^{I}\right)$. The additional GA memory is dominated by storing the population and offspring chromosomes, each with one gene per reachable state, and is bounded by $\mathcal{O}\left(N\,T\,2^I\,Q_{\max}^{I}\right)$. If the GA is evaluated on the preconstructed DTMDP, the DTMDP transition-storage memory must also be available. The expression above corresponds only to the additional memory introduced by the GA search.

\begin{table}[!htb]
\small
\centering
\caption{Simplified worst-case complexity expressed with model and GA parameters.}
\label{tab:simple-complexity-model-parameters}
{%
\begin{tabular}{lll}
\toprule
\bf{Method / step} & \bf{Time complexity} & \bf{Space complexity} \\
\midrule
\bf{DTMDP state bound}
&
--
& $\mathcal{O}\left(T2^I Q_{\max}^{I}\right)$ \\
\midrule
\bf{DTMDP action bound}
&
--
& $\mathcal{O}\left(Q_{\max}^{I}\right)$ \\
\midrule
\bf{DTMDP construction}
&
$\mathcal{O}\left(T2^I Q_{\max}^{2I}(I+n+2^I)\right)$
&
$\mathcal{O}\left(T4^I Q_{\max}^{2I}\right)$ \\
\midrule
\bf{Exact finite-horizon resolution}
&
$\mathcal{O}\left(T^2 4^I Q_{\max}^{2I}\right)$
&
$\mathcal{O}\left(T^2 2^I Q_{\max}^{I}\right)$ \\
\midrule
\bf{GA full run}
&
$\mathcal{O}\left(GNT4^I Q_{\max}^{I}\right)$
&
$\mathcal{O}\left(NT2^I Q_{\max}^{I}\right)$ \\
\bottomrule
\end{tabular}%
}
\end{table}

The main structural difference is the action factor. The exact method must evaluate all admissible actions in every state, which introduces the additional factor $Q_{\max}^I$. In contrast, the GA evaluates fixed policies, where each chromosome already specifies one action per state. This explains why the GA can become useful when the DTMDP state and action spaces grow, even though it does not provide an optimality certificate.

\section{Experimental study}
\label{sec:experiments}

This section evaluates the computational effect of stochastic timing and the performance of the proposed genetic algorithm on a heterogeneous benchmark set. The experiments assess three aspects: the additional complexity introduced by stochastic arrivals relative to a deterministic timing baseline, the solution quality of the GA, and its computational speed and memory scalability. The benchmark combines controlled instance families, in which one structural parameter is varied at a time, and broader heterogeneous instances involving different horizons, quantities, numbers of demands, numbers of products, and arrival-window lengths.

To isolate the effect of stochastic timing, each stochastic instance is paired with a deterministic counterpart. In this counterpart, each arrival distribution is replaced by a single arrival period chosen as the mode of the distribution, while all quantities, deadlines, products, capacities, and cost parameters remain unchanged. The deterministic counterpart is solved by an exact deterministic dynamic program with states of the form \((t,P)\), without arrival-status variables and without stochastic transition matrices. This comparison provides a baseline for measuring the additional state-space growth, transition growth, solution time, and memory pressure caused by stochastic arrivals.

The experiments were run on a workstation with 15Gi of RAM. The processor was Intel(R) Core(TM) i5-8350U CPU @ 1.70GHz, and the experiments were executed using single-core processing. The operating system was Ubuntu 24.04.4 LTS. The memory limit is important for the interpretation of the results, because several large stochastic DTMDP instances can be constructed but cannot be solved exactly, while some even larger candidates cannot be fully constructed due to memory limitations.

\subsection{Benchmark instance set}
\label{sec:benchmark-instance-set}

The benchmark contains 330 stochastic DTMDP instances. These instances are heterogeneous: they vary in planning horizon, demand quantity, number of demands, number of products, arrival windows, and resulting DTMDP size. We report first the deterministic-counterpart statistics, then the corresponding stochastic DTMDP descriptors, exact stochastic solution times, memory indicators, and finally the GA resolution time. This ordering separates the computational burden introduced by stochastic timing from the subsequent evaluation of the GA as an approximate solution method for the stochastic problem.

Table~\ref{tab:benchmark-summary} reports summary statistics for the most relevant benchmark descriptors.

\begin{table}[!htb]
\centering
\caption{Summary statistics of the benchmark instance set.}
\label{tab:benchmark-summary}
\resizebox{1.00\linewidth}{!}{
\begin{tabular}{lrrrrrl}
\hline
\textbf{Descriptor} & \textbf{Count} & \textbf{Min} & \textbf{Max} & \textbf{Mean} & \textbf{Std.} & \textbf{Unit} \\
\hline
Planning horizon $T$ & 330 & 3 & 301 & 34.50 & 58.27 & periods \\
Number of demands $I$ & 330 & 1 & 7 & 2.11 & 1.45 & demands \\
Number of products & 330 & 1 & 6 & 1.37 & 0.80 & products \\
Mean demand quantity & 330 & 1.00 & $1.02\times 10^{3}$ & 23.18 & 76.46 & units \\
Mean arrival-window length & 330 & 3.00 & 152.00 & 18.78 & 29.49 & periods \\
\hline
Deterministic DP states & 330 & 6 & $9.09\times 10^{3}$ & 785.05 & $1.47\times 10^{3}$ & states \\
Deterministic DP arcs & 330 & 7 & $7.87\times 10^{5}$ & $3.42\times 10^{4}$ & $8.98\times 10^{4}$ & arcs \\
Deterministic DP build time & 330 & 0.002 & 39.50 & 1.43 & 3.96 & seconds \\
Deterministic DP exact solve time & 330 & 0.0004 & 0.79 & 0.04 & 0.09 & seconds \\
Deterministic DP total time & 330 & 0.002 & 40.29 & 1.48 & 4.06 & seconds \\
\hline
Number of stochastic DTMDP states & 330 & 11 & $8.21\times 10^{4}$ & $4.44\times 10^{3}$ & $1.07\times 10^{4}$ & states \\
Number of DTMDP actions & 330 & 2 & $1.03\times 10^{3}$ & 62 & 115 & actions \\
Stochastic non-zero transition entries & 330 & 27 & $8.59\times 10^{7}$ & $1.66\times 10^{6}$ & $7.35\times 10^{6}$ & entries \\
Mean feasible actions per state & 330 & 1.25 & 200.88 & 14.61 & 21.93 & actions/state \\
Stochastic DTMDP construction time & 330 & 0.01 & $7.22\times 10^{3}$ & 141.91 & 617.16 & seconds \\
Stochastic exact resolution time & 310 & 0.01 & $1.74\times 10^{3}$ & 60.97 & 199.71 & seconds \\
Stochastic build plus exact solve time & 310 & 0.02 & $7.22\times 10^{3}$ & 199.01 & 660.60 & seconds \\
Stochastic DTMDP build peak RAM & 330 & 9.00 & 54.40 & 18.38 & 8.67 & \% \\
Stochastic exact resolution peak RAM & 310 & 9.00 & 83.50 & 23.37 & 17.22 & \% \\
\hline
Stochastic/deterministic state ratio & 330 & 1.18 & 127.02 & 6.62 & 13.12 & ratio \\
Stochastic/deterministic transition ratio & 330 & 1.83 & $3.57\times 10^{3}$ & 83.51 & 311.40 & ratio \\
Stochastic/deterministic total-time ratio & 310 & 2.00 & $2.36\times 10^{3}$ & 116.32 & 336.81 & ratio \\
\hline
GA resolution time & 330 & 0.02 & $1.11\times 10^{3}$ & 13.04 & 41.08 & seconds \\
\hline
\end{tabular}
}
\end{table}

The mean number of deterministic DP states is 785.05 and the mean deterministic total solution time is 1.48 seconds. This confirms that the underlying deterministic lot-sizing structure can still be handled efficiently by a tailored dynamic program on many instances.

The stochastic DTMDP descriptors show the additional burden created by uncertain arrival timing. The mean number of stochastic states increases to $4.44\times 10^3$, and the largest stochastic transition model contains $8.59\times 10^7$ non-zero entries. On average, the stochastic model has 6.62 times more states and 83.51 times more transition entries than its deterministic counterpart. The mean stochastic build plus exact solve time is 199.01 seconds on the solved instances, and the average stochastic/deterministic total-time ratio is 116.32. These values show that stochastic timing is not a minor extension of deterministic lot sizing; it changes the computational nature of the problem by introducing arrival-status states and probabilistic branching.

From a memory perspective, the stochastic DTMDP approach is also more demanding. The peak RAM during exact resolution reaches 83.50\% on the tested machine, and some stochastic instances cannot be solved exactly because the value and policy tables must be stored in addition to the transition model. The GA resolution-time row is discussed in the subsequent analysis.

\subsection{Experimental setup and performance metrics}
\label{sec:experimental-setup}

The common experimental protocol is organized in two steps. First, the stochastic instance is solved by the exact DTMDP whenever possible. Second, for each stochastic instance that can be solved exactly, we then compare the optimal expected cost with the GA expected cost. For stochastic instances that cannot be solved exactly, the GA can still return a feasible policy, but no certified optimality gap is available.
All times are wall-clock times. For the stochastic DTMDP, construction time is reported separately from optimization time because the construction phase is common to both the exact DTMDP solver and the GA evaluation. The memory indicators correspond to observed peak RAM percentages during stochastic DTMDP construction and exact stochastic resolution.
The quality loss of the GA is measured by the relative optimality gap:
\begin{equation}
\mathrm{Gap}(\%) =
100\times
\frac{C^{\mathrm{GA}}-C^{\mathrm{opt}}}{C^{\mathrm{opt}}},
\label{eq:gap-global}
\end{equation}
where $C^{\mathrm{opt}}$ is the optimal expected cost obtained by the exact DTMDP and $C^{\mathrm{GA}}$ is the expected cost obtained by the GA. In the following experiments, a gap around $5\%$ is considered a reasonable accuracy limit for the heuristic, since the objective is to obtain feasible policies with substantially lower computational effort on instances where exact resolution becomes expensive.

The computational gain is measured by the speedup:
\begin{equation}
\mathrm{Speedup}=
\frac{t^{\mathrm{exact}}_{\mathrm{solve}}}{t^{\mathrm{GA}}_{\mathrm{solve}}}.
\label{eq:speedup-global}
\end{equation}
The speedup excludes the common DTMDP construction step. This convention is used because both the exact stochastic solver and the GA rely on the same constructed DTMDP representation before optimization starts. Consequently, the speedup compares only the optimization phase: exact finite-horizon resolution versus GA search.
For the GA configuration, the population size is $N=100$, the maximum number of generations is $G=100$, the mutation probability is $0.05$, and the mutation gene fraction is $0.10$. At each generation, each parent pair produces exactly four children: two by gene-uniform crossover and two by time-block crossover. Convergence is declared if the best global fitness does not improve by more than $10^{-1}$ for $10$ consecutive generations. The initialization combines random feasible policies, a small number of constructive policies, and relaxed-DTMDP seeds.

%
For exact-unsolved instances, the empirical Bellman-time model introduced later is used only to estimate the missing exact resolution time and to extrapolate the corresponding speedup; no optimality gap is claimed for these cases.

\subsection{Optimality gap analysis}
\label{sec:exp-optimality-gap}

\begin{figure}[!htb]
\centering

\begin{subfigure}{0.48\linewidth}
\centering
\resizebox{\linewidth}{!}{
\begin{tikzpicture}
\begin{axis}[%
grid=both,
width=3in,
height=3in,
xlabel={$C^{\mathrm{opt}}$},
ylabel={$C^{\mathrm{GA}}$},
xmode=log,
ymode=log,
xmin=1,
xmax=2500,
ymin=1,
ymax=2500,
scale only axis,
separate axis lines,
every outer x axis line/.append style={white!15!black},
every x tick label/.append style={font=\color{white!15!black}},
every outer y axis line/.append style={white!15!black},
every y tick label/.append style={font=\color{white!15!black}},
legend style={draw=white!15!black,legend cell align=left,legend pos=south east}
]

\addplot [color=black,only marks,mark=*, mark options={fill=red}]
table[row sep=crcr]{%
2.66431985272132  3.12652426197637    \\
2.66431985272132  3.12652426197637    \\
7.97663326492507  9.01153110682068    \\
16.4783626897996  18.5346390665474    \\
7.97663326492507  8.96932129699928    \\
10.744671895318 11.8851431908729    \\
10.744671895318 11.8516712497007    \\
11.7799444047798  12.943451730654   \\
10.0839556186498  11.0686143354038    \\
19.5212493696008  21.3719379399709    \\
2.66431985272132  2.91099456929186    \\
9.12748759108091  9.96040498275506    \\
17.4668777406875  18.9548759704208    \\
10.0839556186498  10.8815323211486    \\
20.6180791009143  22.2329334831844    \\
35.7948292223806  38.3901177451003    \\
23.9855257021832  25.7228408121612    \\
11.6168249009747  12.455208845284   \\
20.9400445377969  22.4374235761386    \\
168.830927550399  180.220875719619    \\
18.8684567857915  20.1056412870474    \\
13.7546280535731  14.6145872638758    \\
16.3876363963006  17.3999274520304    \\
167.431985272133  177.720600364944    \\
17.6431985272133  18.7204315718424    \\
2.66431985272132  2.8259954011098   \\
2.66431985272132  2.8259954011098   \\
2.66431985272132  2.8259954011098   \\
2.66431985272132  2.8259954011098   \\
2.66431985272132  2.82599540105496    \\
2.66431985272132  2.82558943202107    \\
10.0537814964568  10.6485545405963    \\
17.6431985272133  18.6637974301341    \\
17.8893390895773  18.9158952119126    \\
167.431985272133  176.590465716587    \\
167.431985272133  176.272139114438    \\
9.12748759108091  9.60407685269531    \\
167.431985272133  175.774693126913    \\
167.431985272133  175.761422625825    \\
167.431985272133  175.753926160147    \\
167.431985272133  175.745933103678    \\
7.50594330788897  7.8780748470494   \\
167.431985272133  175.728264090974    \\
167.431985272133  175.713930277622    \\
52.8002690091526  55.4048437100795    \\
4.19743437138373  4.40366616101841    \\
4.19743437138373  4.40366616101841    \\
168.830927550399  177.050228820446    \\
7.50594330788897  7.8700646705226   \\
20.4426343336271  21.4307721246756    \\
423.878396839781  443.989932886351    \\
50.9295955816397  53.3370650396927    \\
53.8597996049727  56.3976435247709    \\
168.830927550399  176.764113897378    \\
106.719599209945  111.713424888358    \\
11.141527723162 11.6621500601337    \\
17.6431985272133  18.4669705953335    \\
17.6431985272129  18.4669705953328    \\
167.431985272133  175.235511975106    \\
50.9295955816386  53.2785069020485    \\
14.2149499012432  14.8668186388986    \\
27.8696337598702  29.1431301301684    \\
167.431985272133  175.058066988693    \\
5.64441171436718  5.90075803725185    \\
50.9295955816397  53.2309134496364    \\
50.9295955816397  53.2252373730742    \\
212.439198419891  221.985798581837    \\
168.830927550399  176.384375735031    \\
167.431985272133  174.838923315541    \\
50.6025667666766  52.8402045294033    \\
50.5560621296619  52.7881027192569    \\
167.431985272133  174.818545825081    \\
17.6431985272133  18.4161506303474    \\
5.64441171436718  5.88923053404498    \\
25.7780310648309  26.8920558826881    \\
133.149499012432  138.892056531371    \\
498.570759613816  519.717768557596    \\
8.74068619061197  9.10940794222871    \\
15.4452651984728  16.0934341132514    \\
5.96008300944696  6.20803445892827    \\
106.719599209945  111.012114914911    \\
5.41954768549593  5.63512742884759    \\
10.8567237413827  11.2829421865545    \\
7.50594330788897  7.79962150208351    \\
2.66431985272132  2.7674372634677   \\
2.66431985272132  2.7674372634677   \\
2.66431985272132  2.7674372634677   \\
2.66431985266035  2.76743278498587    \\
167.431985272133  173.758198370951    \\
167.431985272133  173.758198370951    \\
167.431981820182  173.758191952566    \\
50.9295955816397  52.8453824349787    \\
9.20485500372733  9.55062781676039    \\
168.781190211697  175.081444661867    \\
50.9146419796134  52.8007923488005    \\
2.66431985272132  2.76287009204989    \\
2.66431985272132  2.76273476902031    \\
167.382139932045  173.559085659912    \\
5.86552790083817  6.08172084325543    \\
167.431985272129  173.567289074774    \\
74.9863447482738  77.7322913232143    \\
168.830927550396  174.927375241296    \\
167.382139932045  173.396734333738    \\
17.6382139932045  18.2713638311087    \\
17.6431985272133  18.2758198370951    \\
23.0384856696986  23.845288900109   \\
5.96008300944696  6.16868144006251    \\
5.86552790083817  6.06984640929153    \\
14.1083577277601  14.5940514960262    \\
16.7417754783561  17.3180635000779    \\
67.0747495062158  69.3804353236048    \\
48.2253264350684  49.8645815861462    \\
17.6431985272133  18.2407372074902    \\
225.115176879824  232.641869407229    \\
7.50594330788897  7.75530807627287    \\
50.9295955798104  52.6035160269992    \\
100.112124259324  103.381445332302    \\
27.4298998024863  28.3017187552806    \\
27.4298998024863  28.3017187552806    \\
5.86552790083817  6.05052636587373    \\
17.518687376554 18.0635742220504    \\
5.86552790083817  6.04630973719616    \\
14.2149499012432  14.6508593776403    \\
53.8597996049727  55.4964788614921    \\
1254.50690387996  1292.50690387996    \\
5.72253264350684  5.89541905002336    \\
158.417754783561  163.166879360612    \\
9.25934368827698  9.53178711102519    \\
12.5085924790381  12.869950883238   \\
7.60747495062158  7.82542968882016    \\
7.60747495062158  7.82542968882016    \\
5.95560621296619  6.11907226661512    \\
12.2983839514885  12.6318063500078    \\
933.514842368037  958.709858973834    \\
167.431985272133  171.704780823358    \\
50.9295955186237  52.1852135329894    \\
167.431985272129  171.504892560837    \\
2.66431985272132  2.72758198370951    \\
2.66431985272132  2.72758198370951    \\
2.66431985272132  2.72758198370951    \\
2.66431985272132  2.72758198370951    \\
2.66431985272132  2.72758198370951    \\
2.66431985272132  2.72758198370951    \\
2.66431985272132  2.72758198370951    \\
2.66431985272132  2.72758198370951    \\
2.66431985272132  2.72758198370951    \\
2.66431985272129  2.72758198370944    \\
2.66431985272129  2.72758198370944    \\
2.66431985062079  2.72758195156025    \\
9.12748759108091  9.33964850875359    \\
2.57417754783561  2.63180635000779    \\
2.66431985272132  2.7184476408739   \\
2.66431985272132  2.71844762261884    \\
2.66431985272132  2.71844762261884    \\
2.66431985272132  2.71844762261883    \\
2.66431985272132  2.71831231784433    \\
4.19743437138373  4.27326829513377    \\
168.830927550399  171.740890940188    \\
159.901206037898  162.620510828107    \\
17.6431985062079  17.9244305826919    \\
4.27559141365382  4.34130388983536    \\
12.4358448456052  12.6181516049893    \\
34.0373747531079  34.4732842295051    \\
167.431985062079  169.550416344504    \\
2156.36933596068  2183.36933596068    \\
4.19743437138373  4.23685805197805    \\
4.27559141365382  4.31001771055586    \\
8.97427473822294  9.04518785575432    \\
168.830927550399  169.359885022649    \\
168.830927336701  169.359582053824    \\
2.66431985272132  2.6690238460674   \\
2.66431985272132  2.6690238460674   \\
2.66431985272132  2.6690238460674   \\
2.66431985272132  2.6690238460674   \\
2.66431985272132  2.6690238460674   \\
2.66431985272132  2.6690238460674   \\
2.66431985272132  2.6690238460674   \\
2.66431985272132  2.6690238460674   \\
2.66431985272132  2.6690238460674   \\
2.66431985272132  2.6690238460674   \\
2.66431985272132  2.6690238460217   \\
2.66431985272132  2.66902382172731    \\
2.66431985272132  2.66901936765382    \\
2.66431985272132  2.66901936765382    \\
2.66431985272132  2.66875320000825    \\
2.66431985272132  2.66875170110955    \\
2.66431985272132  2.6644566746496   \\
2.66431985272132  2.66445667464957    \\
2.66431985272132  2.66445368904054    \\
2.66431985272132  2.66445368904054    \\
2.66431985272132  2.66432133944998    \\
2.54813723812239  2.54813723812239    \\
2.54813723812239  2.54813723812239    \\
2.57417754783561  2.57417754783561    \\
2.63180635000779  2.63180635000779    \\
2.57417754783561  2.57417754783561    \\
2.6518687376554 2.6518687376554   \\
2.57417754783561  2.57417754783561    \\
2.54813723812239  2.54813723812239    \\
2.65341889222255  2.65341889222255    \\
2.6518687376554 2.6518687376554   \\
2.57417754783561  2.57417754783561    \\
4.14835509567122  4.14835509567122    \\
4.14835509567122  4.14835509567122    \\
4.30373747531079  4.30373747531079    \\
2.57417754783561  2.57417754783561    \\
2.57417754783561  2.57417754783561    \\
2.66382139932045  2.66382139932045    \\
2.66382139932045  2.66382139932045    \\
4.30373747531079  4.30373747531079    \\
4.09627447624479  4.09627447624479    \\
4.09627447624479  4.09627447624479    \\
8.87088773917806  8.87088773917806    \\
8.41364155583282  8.41364155583282    \\
7.19254895248958  7.19254895248958    \\
7.19254895248958  7.19254895248958    \\
8.74068619061197  8.74068619061197    \\
17.1831263471441  17.1831263471441    \\
10.2888234287344  10.2888234287344    \\
10.2888234287344  10.2888234287344    \\
};
\addlegendentry{GA policies};

\addplot [color=blue,ultra thick,dashdotted]
table[row sep=crcr]{%
2    2  \\
2200   2200  \\
};
\addlegendentry{Reference line \(y=x\)};

\end{axis}
\end{tikzpicture}
}
\caption{All tested instances}
\label{fig:ga-cost-vs-optimal-cost-all}
\end{subfigure}
\hfill
\begin{subfigure}{0.48\linewidth}
\centering
\resizebox{\linewidth}{!}{
\begin{tikzpicture}
\begin{axis}[%
grid=both,
width=3in,
height=3in,
xlabel={$C^{\mathrm{opt}}$},
ylabel={$C^{\mathrm{GA}}$},
xmin=0,
xmax=40,
ymin=0,
ymax=40,
scale only axis,
separate axis lines,
every outer x axis line/.append style={white!15!black},
every x tick label/.append style={font=\color{white!15!black}},
every outer y axis line/.append style={white!15!black},
every y tick label/.append style={font=\color{white!15!black}},
legend style={draw=white!15!black,legend cell align=left,legend pos=south east}
]

\addplot [color=black,only marks,mark=*, mark options={fill=red}]
table[row sep=crcr]{%
35.7948292223806  38.3901177451003  \\
23.9855257021832  25.7228408121612  \\
20.9400445377969  22.4374235761386  \\
20.6180791009143  22.2329334831844  \\
19.5212493696008  21.3719379399709  \\
18.8684567857915  20.1056412870474  \\
17.4668777406875  18.9548759704208  \\
17.8893390895773  18.9158952119126  \\
17.6431985272133  18.7204315718424  \\
17.6431985272133  18.6637974301341  \\
16.4783626897996  18.5346390665474  \\
16.3876363963006  17.3999274520304  \\
13.7546280535731  14.6145872638758  \\
11.7799444047798  12.943451730654 \\
11.6168249009747  12.455208845284 \\
10.744671895318 11.8851431908729  \\
10.744671895318 11.8516712497007  \\
10.0839556186498  11.0686143354038  \\
10.0839556186498  10.8815323211486  \\
10.0537814964568  10.6485545405963  \\
9.12748759108091  9.96040498275506  \\
9.12748759108091  9.60407685269531  \\
7.97663326492507  9.01153110682068  \\
7.97663326492507  8.96932129699928  \\
2.66431985272132  3.12652426197637  \\
2.66431985272132  3.12652426197637  \\
2.66431985272132  2.91099456929186  \\
2.66431985272132  2.8259954011098 \\
2.66431985272132  2.8259954011098 \\
2.66431985272132  2.8259954011098 \\
2.66431985272132  2.8259954011098 \\
2.66431985272132  2.82599540105496  \\
2.66431985272132  2.82558943202107  \\
};
\addlegendentry{GA policies};

\addplot [color=blue,ultra thick,dashdotted]
table[row sep=crcr]{%
2    2  \\
40  40  \\
};
\addlegendentry{Reference line \(y=x\)};

\end{axis}
\end{tikzpicture}
}
\caption{Worst-gap instances}
\label{fig:ga-cost-vs-optimal-cost-worst}
\end{subfigure}

\caption{Comparison between the optimal expected cost and the GA expected cost. The left panel reports all tested instances, while the right panel focuses on the instances with the largest observed optimality gaps.}
\label{fig:ga-cost-vs-optimal-cost-subfigures}
\end{figure}
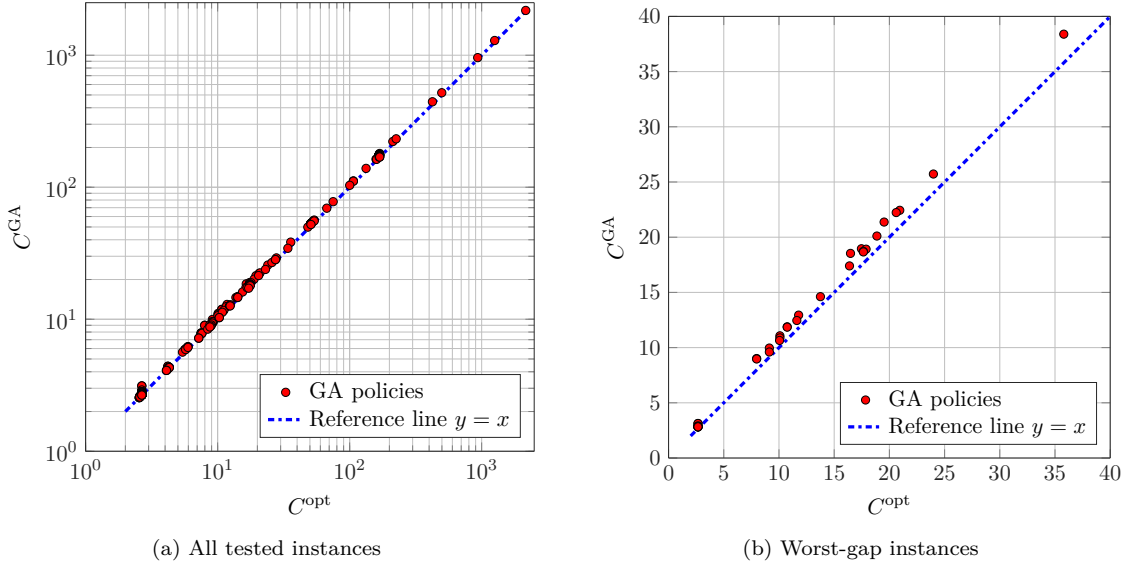

This section evaluates the solution quality of the genetic algorithm over the instances for which an exact DTMDP reference cost is available. Over the complete set of solved instances, the average optimality gap is $3.44\% \pm 0.39\%$ within a 95\% confidence interval, with a standard deviation of 2.95 percentage points. This indicates that the GA remains close to the exact optimum on average while maintaining moderate dispersion across heterogeneous instances. Most importantly, the average gap remains below the 5\% accuracy threshold used as a practical reference in this study. On the difficult benchmark instances, comprising 90 test cases, we still satisfy this 5\% quality limit, and the GA achieves an average optimization speedup of \(6.89\pm1.41\) at the 95\% confidence level. This result is important because it shows that the GA acceleration is not obtained by allowing arbitrary degradation of solution quality; it is measured under a controlled optimality-gap requirement.
Figure~\ref{fig:ga-cost-vs-optimal-cost-all} compares the GA cost with the optimal cost over all tested instances. This type of parity plot is commonly used to assess whether two numerical quantities are close to each other. The dashed line corresponds to the ideal case \(y=x\), where the GA would return exactly the same cost as the exact DTMDP method.

The point cloud in Figure~\ref{fig:ga-cost-vs-optimal-cost-all} is almost aligned with the reference line \(y=x\). This indicates a strong agreement between the GA costs and the exact optimal costs. Even for instances with larger optimal costs, the points remain close to the diagonal, showing that the approximation quality is preserved across different cost magnitudes.
To examine the most difficult cases more closely, Figure~\ref{fig:ga-cost-vs-optimal-cost-worst} reports only the subset of instances with the largest observed gaps. This second figure removes the visual dominance of the full data set and focuses on the cases where the GA deviates the most from the exact solution.
Figure~\ref{fig:ga-cost-vs-optimal-cost-worst} confirms that the largest deviations remain relatively small in absolute terms. Although these instances correspond to the worst observed gaps, the points are still located close to the reference line. This suggests that the GA does not fail catastrophically on difficult cases; instead, the loss of optimality remains controlled.


\subsection{Impact of the planning horizon on GA performance}
\label{sec:exp-planning-horizon}

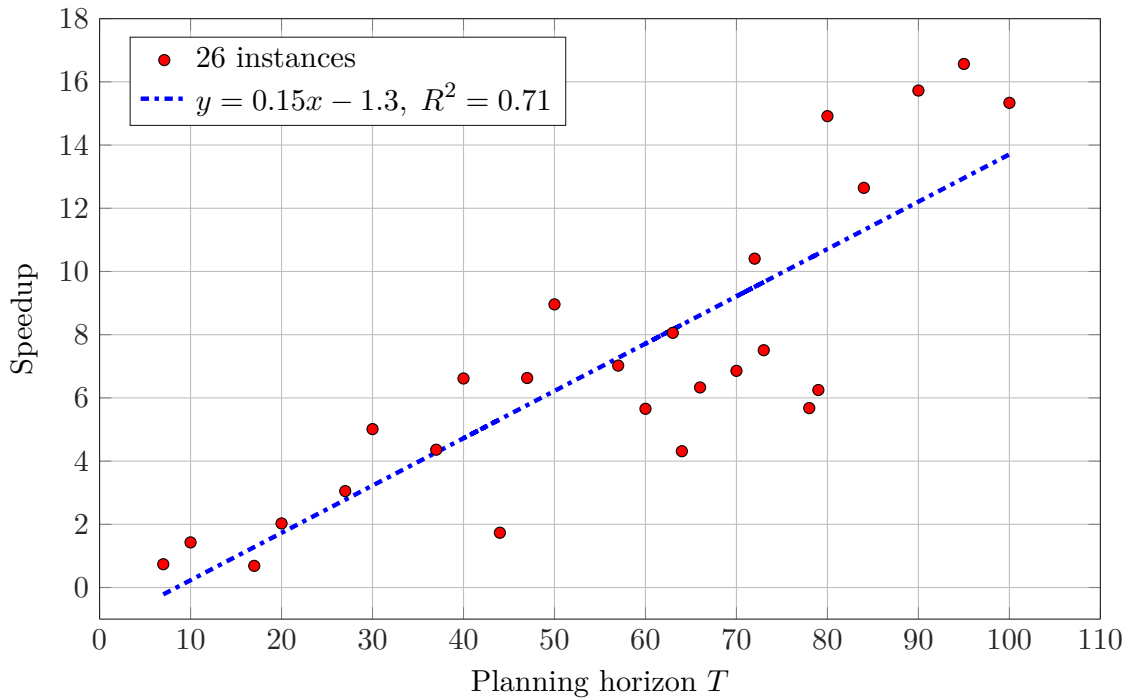
\begin{figure}[!htb]
\centering
\resizebox{1.00\linewidth}{!}{
\begin{tikzpicture}
\begin{axis}[%
grid=both,
width=5in,
height=3in,
at={(0in,0in)},
xlabel={Planning horizon $T$},
ylabel={Speedup},
scale only axis,
separate axis lines,
every outer x axis line/.append style={white!15!black},
every x tick label/.append style={font=\color{white!15!black}},
xmin=0,
xmax=110,
every outer y axis line/.append style={white!15!black},
every y tick label/.append style={font=\color{white!15!black}},
ymin=-1,
ymax=18,
legend style={draw=white!15!black,legend cell align=left,legend pos=north west}
]
\addplot [color=black,only marks,mark=*, mark options={fill=red}]
  table[row sep=crcr]{%
7   0.7368  \\
10    1.4277  \\
17    0.6854  \\
20    2.0279  \\
27    3.0507  \\
30    5.0122  \\
37    4.3603  \\
44    1.7334  \\
40    6.614 \\
47    6.6294  \\
50    8.9587  \\
57    7.022 \\
64    4.3129  \\
60    5.655 \\
63    8.0589  \\
66    6.3308  \\
73    7.5087  \\
70    6.8557  \\
72    10.406  \\
79    6.2491  \\
78    5.6773  \\
80    14.9146 \\
84    12.6458 \\
95    16.5655 \\
90    15.7249 \\
100   15.3354 \\
};
\addlegendentry{26 instances};

%

\addplot [color=blue,ultra thick,dashdotted]
  table[row sep=crcr]{%
7       -0.2171 \\
10        0.232 \\
17        1.2799  \\
20        1.729 \\
27        2.7769  \\
30        3.226 \\
37        4.2739  \\
44        5.3218  \\
40        4.723 \\
47        5.7709  \\
50        6.22  \\
57        7.2679  \\
64        8.3158  \\
60        7.717 \\
66        8.6152  \\
73        9.6631  \\
70        9.214 \\
72        9.5134  \\
79        10.5613 \\
78        10.4116 \\
80        10.711  \\
84        11.3098 \\
90        12.208  \\
100       13.705  \\
};
\addlegendentry{$y=0.15x-1.3,\;R^2=0.71$};
\end{axis}
\end{tikzpicture}%
}
\caption{Speedup of the genetic algorithm as a function of the planning horizon. The speedup excludes the common DTMDP construction time and compares only the optimization phase of the exact method and the GA.}
\label{fig:planning-horizon-speedup}
\end{figure}

We next analyze the effect of the planning horizon on the performance of the genetic algorithm. To isolate this effect, we consider a set of single-demand instances with one product and a fixed demand quantity equal to $Q=100$. The planning horizon $T$ is progressively increased from $7$ to $70$ periods. As the horizon increases, the admissible arrival window is also enlarged. The arrival distribution remains unimodal in all instances, with its peak located at the centre of the corresponding arrival window.
This experimental setting is designed to study how the genetic algorithm behaves when the temporal dimension of the decision problem increases, while keeping the number of products and demands fixed. Hence, the observed changes in computation time and solution quality are mainly driven by the length of the planning horizon.
Figure~\ref{fig:planning-horizon-speedup} reports the speedup obtained by the genetic algorithm with respect to the exact dynamic programming method.

The results show that the exact method remains more efficient for the smallest horizons. In these cases, the DTMDP can still be solved quickly by the exact finite-horizon procedure, whereas the genetic algorithm requires a fixed computational effort for population initialization, chromosome evaluation, crossover, mutation, and selection. 

When the planning horizon increases, the behaviour changes. The genetic algorithm becomes faster than the exact method for medium and larger horizons, and the speedup follows an overall increasing trend. This indicates that the GA becomes more attractive when the temporal dimension expands and the exact solution procedure becomes more expensive. The figure also shows that the gain is not only marginal: for the largest horizons, the GA provides a clear reduction in solve time compared with the exact method.

In terms of solution quality, the genetic algorithm remains close to the exact optimum over this set of instances. The average optimality gap is approximately $5\%$. This indicates that, despite the increase in the planning horizon, the GA is able to preserve a moderate loss of optimality while significantly reducing the optimization time for the larger instances. The results therefore suggest that the proposed GA provides a relevant compromise between computational efficiency and solution quality when the planning horizon becomes large. From a memory perspective, increasing the horizon also enlarges the transition model because more time-indexed states and state-action transitions must be stored. This reinforces the motivation for using the GA when the exact finite-horizon recursion becomes costly on the constructed DTMDP.


\subsection{Impact of the demand quantity on GA performance}
\label{sec:exp-demand-quantity}

\begin{figure}[!htb]
\centering
\resizebox{1.00\linewidth}{!}{
\begin{tikzpicture}
\begin{axis}[%
grid=both,
width=5in,
height=3in,
xlabel={Demand quantity \(Q\)},
ylabel={Speedup},
xmode=log,
scale only axis,
separate axis lines,
every outer x axis line/.append style={white!15!black},
xmin=0.8,
xmax=1300,
every outer y axis line/.append style={white!15!black},
every y tick label/.append style={font=\color{white!15!black}},
ymin=-1,
ymax=6,
legend style={draw=white!15!black,legend cell align=left,legend pos=north west}
]

\addplot [color=black,only marks,mark=*, mark options={fill=red}]
table[row sep=crcr]{%
1     0.2942 \\
2     0.1383 \\
4     0.1237 \\
8     0.0806 \\
16    0.0731 \\
32    0.1464 \\
64    0.4688 \\
128   1.5318 \\
256   4.0189 \\
512   4.2492 \\
1024  5.1539 \\
};
\addlegendentry{11 instances};

\addplot [color=blue,ultra thick,dashdotted]
table[row sep=crcr]{%
32					-0.181297	\\
64					0.929154	\\
128					2.039605	\\
256					3.150056	\\
512					4.260507	\\
1024					5.370958	\\
};
\addlegendentry{$y=1.11\log_2(Q)-5.73,\;R^2=0.92$};

\end{axis}
\end{tikzpicture}%
}
\caption{Speedup of the genetic algorithm as a function of the demand quantity. The speedup excludes the common DTMDP construction time and compares only the optimization phase of the exact method and the GA.}
\label{fig:demand-quantity-speedup}
\end{figure}
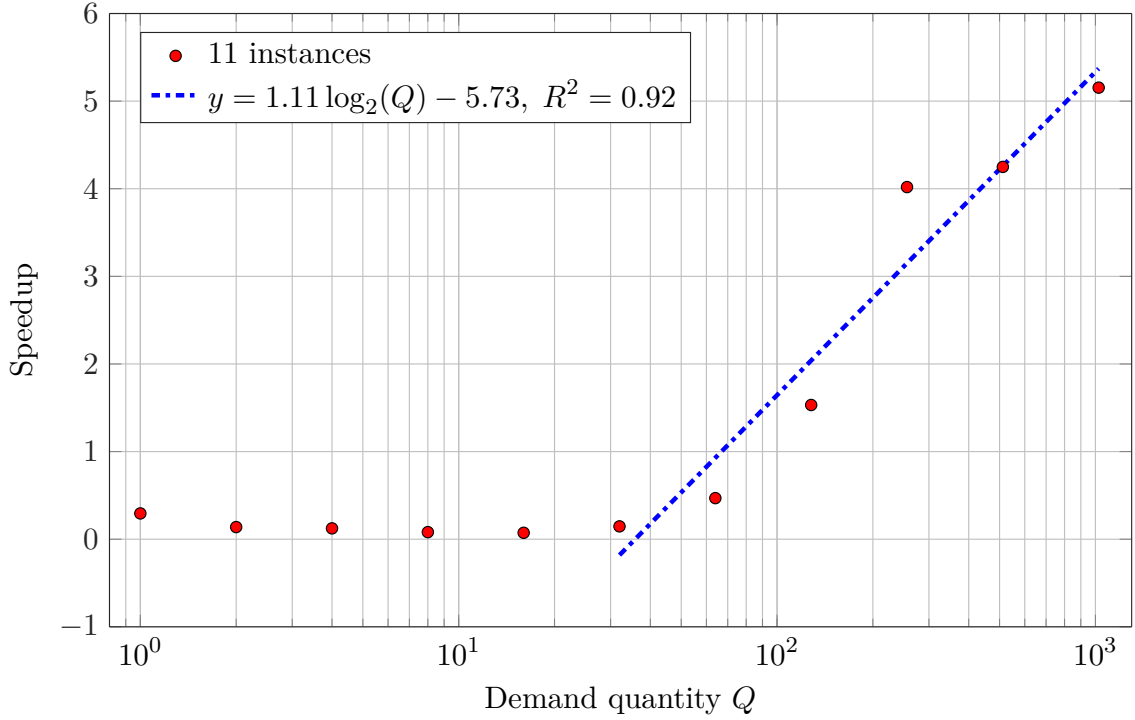

We now analyze the effect of the demand quantity on the performance of the genetic algorithm. In this experiment, we consider single-demand instances with one product, while progressively increasing the demand quantity \(Q\) from \(1\) to \(1024\). The other experimental settings are kept fixed in order to isolate the impact of the quantity level.
Increasing the demand quantity directly affects the size of the decision problem. Indeed, larger quantities increase the number of possible cumulative production levels and enlarge the feasible production allocation space. As a result, the exact DTMDP solution becomes increasingly expensive when \(Q\) grows.
Figure~\ref{fig:demand-quantity-speedup} reports the speedup obtained by the genetic algorithm with respect to the exact DTMDP method as a function of the demand quantity.

The results show that the exact method remains more efficient for small demand quantities. For \(Q\leq 64\), the speedup is smaller than one, meaning that the GA is slower than the exact method. This is expected because, for small quantities, the exact DTMDP can still be solved quickly, while the GA requires a fixed computational effort for initialization, fitness evaluation, crossover, mutation, and selection.

When the demand quantity increases, the behaviour changes. Starting from \(Q=128\), the GA becomes faster than the exact method. The speedup then increases substantially for larger quantities and reaches approximately \(5.15\) for \(Q=1024\). This confirms that the GA becomes more attractive when the quantity dimension expands and the exact finite-horizon resolution becomes more expensive.

In terms of solution quality, the average optimality gap over this set of instances is approximately \(3.52\%\). The GA recovers the exact optimum for the smallest quantity and remains close to the exact solution for larger quantities. The gap stays moderate across the tested range, which indicates that increasing the demand quantity mainly affects computational efficiency rather than causing a strong degradation in solution quality. Larger quantities also increase the number of cumulative production levels, which enlarges both the state space and the action catalogue. This affects memory consumption before it affects only solve time, because the exact method must store the corresponding feasible actions, rewards, and transitions.


\subsection{Impact of the number of demands on GA performance}
\label{sec:exp-number-demands}

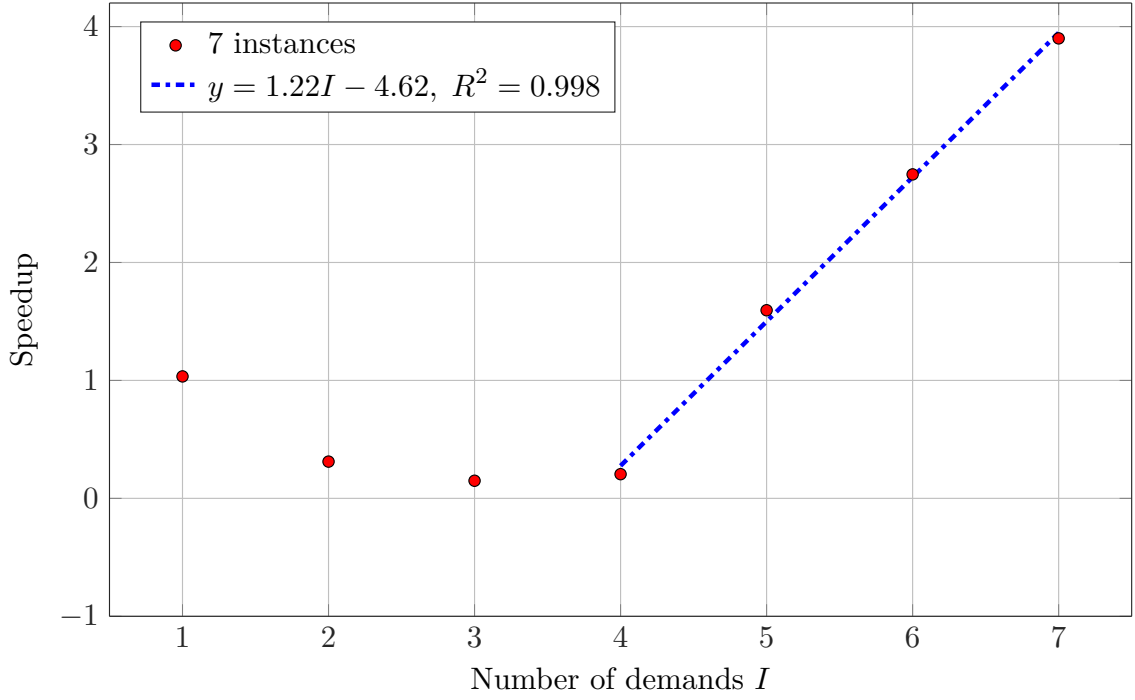
\begin{figure}[!htb]
\centering
\resizebox{1.00\linewidth}{!}{
\begin{tikzpicture}
\begin{axis}[%
grid=both,
width=5in,
height=3in,
xlabel={Number of demands \(I\)},
ylabel={Speedup},
scale only axis,
separate axis lines,
every outer x axis line/.append style={white!15!black},
every x tick label/.append style={font=\color{white!15!black}},
xmin=0.5,
xmax=7.5,
xtick={1,2,3,4,5,6,7},
every outer y axis line/.append style={white!15!black},
every y tick label/.append style={font=\color{white!15!black}},
ymin=-1,
ymax=4.2,
legend style={draw=white!15!black,legend cell align=left,legend pos=north west}
]

\addplot [color=black,only marks,mark=*, mark options={fill=red}]
table[row sep=crcr]{%
1  1.0339 \\
2  0.3116 \\
3  0.1492 \\
4  0.2051 \\
5  1.5949 \\
6  2.7468 \\
7  3.9000 \\
};
\addlegendentry{7 instances};

\addplot [color=blue,ultra thick,dashdotted]
table[row sep=crcr]{%
4  0.2762 \\
5  1.4999 \\
6  2.7235 \\
7  3.9472 \\
};
\addlegendentry{$y=1.22I-4.62,\;R^2=0.998$};

\end{axis}
\end{tikzpicture}%
}
\caption{Speedup of the genetic algorithm as a function of the number of demands. The speedup excludes the common DTMDP construction time and compares only the optimization phase of the exact method and the GA.}
\label{fig:number-demands-speedup}
\end{figure}

We now analyze the effect of the number of demands on the performance of the genetic algorithm. In this experiment, the number of demand components is progressively increased from \(1\) to \(7\), while the other experimental settings are kept fixed.
Increasing the number of demands directly increases the complexity of the decision process. Indeed, each additional demand introduces a new arrival indicator and a new cumulative production component in the state description. It also increases the number of feasible production allocations that must be considered. Therefore, the exact DTMDP solution becomes increasingly expensive as the number of demands grows.
Figure~\ref{fig:number-demands-speedup} reports the speedup obtained by the genetic algorithm with respect to the exact DTMDP method as a function of the number of demands.

The results show that the exact method remains more efficient for a small number of demands. For \(I=2\), \(I=3\), and \(I=4\), the speedup is smaller than one, meaning that the GA is slower than the exact method. This is consistent with the previous experiments: when the instance is still small, the exact DTMDP can be solved quickly, whereas the GA requires a fixed computational effort for initialization, fitness evaluation, crossover, mutation, and selection.

When the number of demands increases, the behaviour changes. Starting from \(I=5\), the GA becomes faster than the exact method, and the speedup increases sharply for larger instances. The speedup reaches approximately \(4\) for \(I=7\). This confirms that the GA becomes more attractive when the number of demand components increases and the exact finite-horizon resolution becomes more expensive.

In terms of solution quality, the average optimality gap over this set of instances is approximately \(8.25\%\). The GA recovers the exact optimum for the single-demand case, while the gap increases for larger demand sets but remains moderate overall. The non-monotonic behaviour of the gap suggests that instance difficulty is not determined only by the number of demands, but also by the interaction between demand windows, quantities, probability distributions, capacity constraints, and setup decisions.

For instances with more than \(7\) demands, the exact DTMDP resolution could not be completed in our computational environment because of memory limitations. In particular, on a machine with \(15\) GB of RAM, the exact method starts to run out of memory due to the rapid growth of the state space, the feasible action sets, and the transition matrices. Consequently, no exact reference cost is available for these larger instances, and the optimality gap cannot be computed. However, the GA does not face the same memory bottleneck and remains able to return feasible solutions with a much lower memory footprint and reasonable computation times, typically between \(30\) and \(60\) minutes.

\subsection{Empirical estimation of exact Bellman resolution time}
\label{sec:exp-exact-time-regression}

The previous experiments show that the exact DTMDP approach becomes limited by both computation time and memory consumption when the generated state space and transition matrices become large. In several large instances, the DTMDP can still be constructed, so that the number of states and actions is known, but the exact finite-horizon Bellman resolution cannot be completed on the test machine because of memory limitations. For these cases, the optimal cost is not directly available, and the exact resolution time cannot be measured.
To still quantify the computational advantage of the genetic algorithm on these large instances, we estimate the missing exact Bellman resolution time using an empirical regression model fitted on the instances that were solved exactly. The objective of this regression is to extrapolate the resolution time that the exact method would require on large instances for which the optimal solution cannot be computed on the available hardware. This estimated exact solve time is then used to approximate the speedup of the genetic algorithm.
Let \(S\) denote the number of reachable DTMDP states and \(A\) the number of actions in the constructed DTMDP. Based on the empirical regression, the exact resolution time is well approximated by the following power law:
\begin{equation}
t_{\mathrm{exact}}
\approx
2.8\times 10^{-8} S^2 A .
\label{eq:empirical-exact-time-model}
\end{equation}
Equivalently, in logarithmic form,
$
\ln(t_{\mathrm{exact}})
\approx
\ln(2.8\times 10^{-8})
+
\ln(S^2 A).
$
The regression is therefore interpreted as a one-variable log--log model, where the explanatory variable is the generated-model size indicator \(S^2A\).

This empirical model is consistent with the complexity discussion in Section~\ref{sec:complexity}. The theoretical analysis shows that exact DTMDP resolution is driven by the generated state space, the action space, and the transition structure. In an ideal sparse Bellman implementation, the cost can be related to the number of state--action pairs and their non-zero successors. However, the exact solver used in the experiments manipulates action-indexed transition matrices over the generated state space. Consequently, for large instances, the observed implementation-level cost is better captured by a matrix-size proxy of order \(S^2A\). The regression in Equation~\eqref{eq:empirical-exact-time-model} should therefore be interpreted as an empirical complexity law for the implemented exact solver on the test machine, rather than as a general theoretical upper bound.

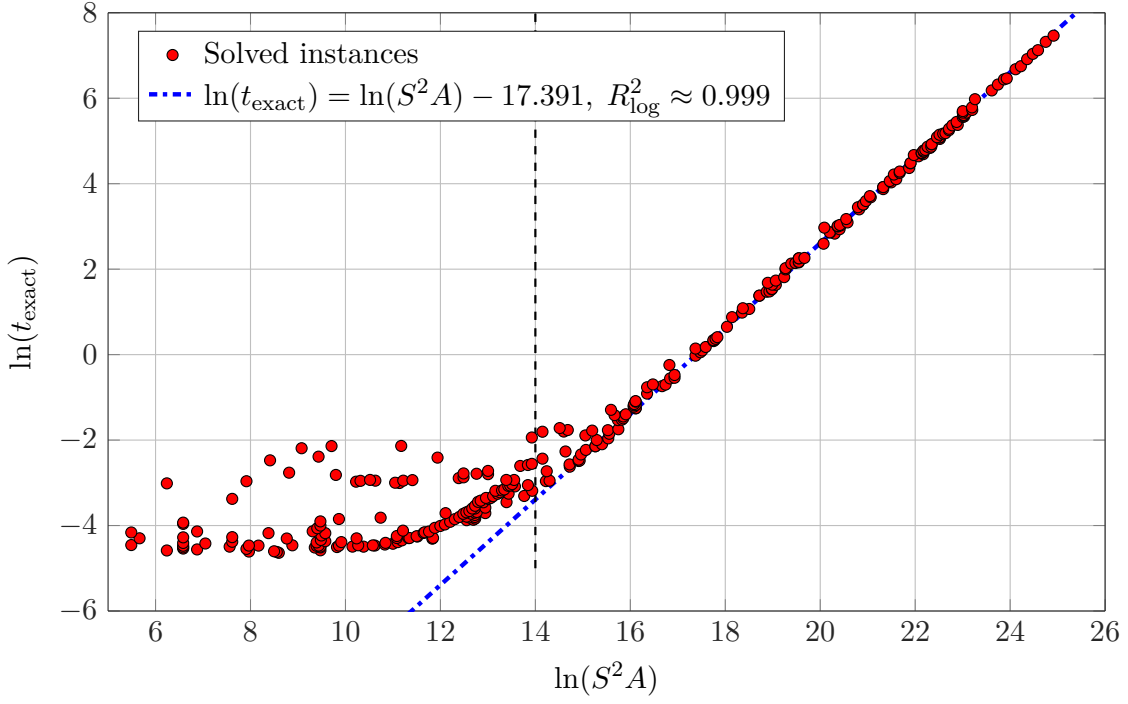
\begin{figure}[!htb]
\centering
\resizebox{1.00\linewidth}{!}{
\begin{tikzpicture}
\begin{axis}[%
grid=both,
width=5in,
height=3in,
xlabel={\(\ln(S^2A)\)},
ylabel={\(\ln(t_{\mathrm{exact}})\)},
scale only axis,
separate axis lines,
every outer x axis line/.append style={white!15!black},
every x tick label/.append style={font=\color{white!15!black}},
every outer y axis line/.append style={white!15!black},
every y tick label/.append style={font=\color{white!15!black}},
xmin=5,
xmax=26,
ymin=-6,
ymax=8,
legend style={draw=white!15!black,legend cell align=left,legend pos=north west}
]

\addplot [
color=black,
only marks,
mark=*,
mark options={solid,fill=red},
unbounded coords=discard
]
table[
col sep=space,
x expr={2*ln(\thisrow{mdp_nb_states}) + ln(\thisrow{mdp_nb_actions})},
y expr={ln(\thisrow{exact_time_sec})}
]{empiric1.csv};
\addlegendentry{Solved instances};

\addplot [
color=blue,
ultra thick,
dashdotted,
domain=5:26,
samples=2
]
{x - 17.3911};
\addlegendentry{\(\ln(t_{\mathrm{exact}})=\ln(S^2A)-17.391,\;R^2_{\log}\approx 0.999\)};

\addplot [
color=black,
dashed,
thick
]
table[row sep=crcr]{%
14  -5 \\
14   8 \\
};

\end{axis}
\end{tikzpicture}%
}
\caption{Empirical regression of the exact Bellman resolution time as a function of the generated DTMDP size indicator \(S^2A\). Both axes are represented in natural logarithmic scale. The fitted line corresponds to \(t_{\mathrm{exact}}\approx 2.8\times 10^{-8}S^2A\). The model achieves \(R^2_{\log}\approx 0.999\) on the logarithmic scale and a mean absolute percentage error of approximately \(5\%\) on the retained large instances.}
\label{fig:exact-time-regression-s2a}
\end{figure}

Figure~\ref{fig:exact-time-regression-s2a} reports the regression in logarithmic scale. Each point corresponds to one benchmark instance for which the exact Bellman resolution time was observed. The horizontal axis is \(\ln(S^2A)\), and the vertical axis is \(\ln(t_{\mathrm{exact}})\). The fitted line corresponds to Equation~\eqref{eq:empirical-exact-time-model}.

The quality of the fit confirms that \(S^2A\) is a strong predictor of the exact Bellman resolution time for large generated DTMDPs. The coefficient of determination in logarithmic scale is \(R^2_{\log}\approx 0.999\), which means that the regression explains almost all the variability of \(\ln(t_{\mathrm{exact}})\) on the fitted instances. The mean absolute percentage error is approximately \(5\%\), indicating that the predicted exact solve time is close to the observed time in relative terms.

This regression is then used only for instances where the exact Bellman resolution cannot be completed because of the memory limit, while the DTMDP construction provides the values of \(S\) and \(A\). For such an instance, the estimated speedup is computed by replacing the missing exact resolution time with the prediction \(\widehat{t}_{\mathrm{exact}}\) obtained from Equation~\eqref{eq:empirical-exact-time-model}. This allows the genetic algorithm to be compared with an estimated exact baseline even when the optimal policy cannot be computed on the available machine.

\subsection{Relationship between experimental complexity and speedup}
\label{sec:exp-complexity-speedup}

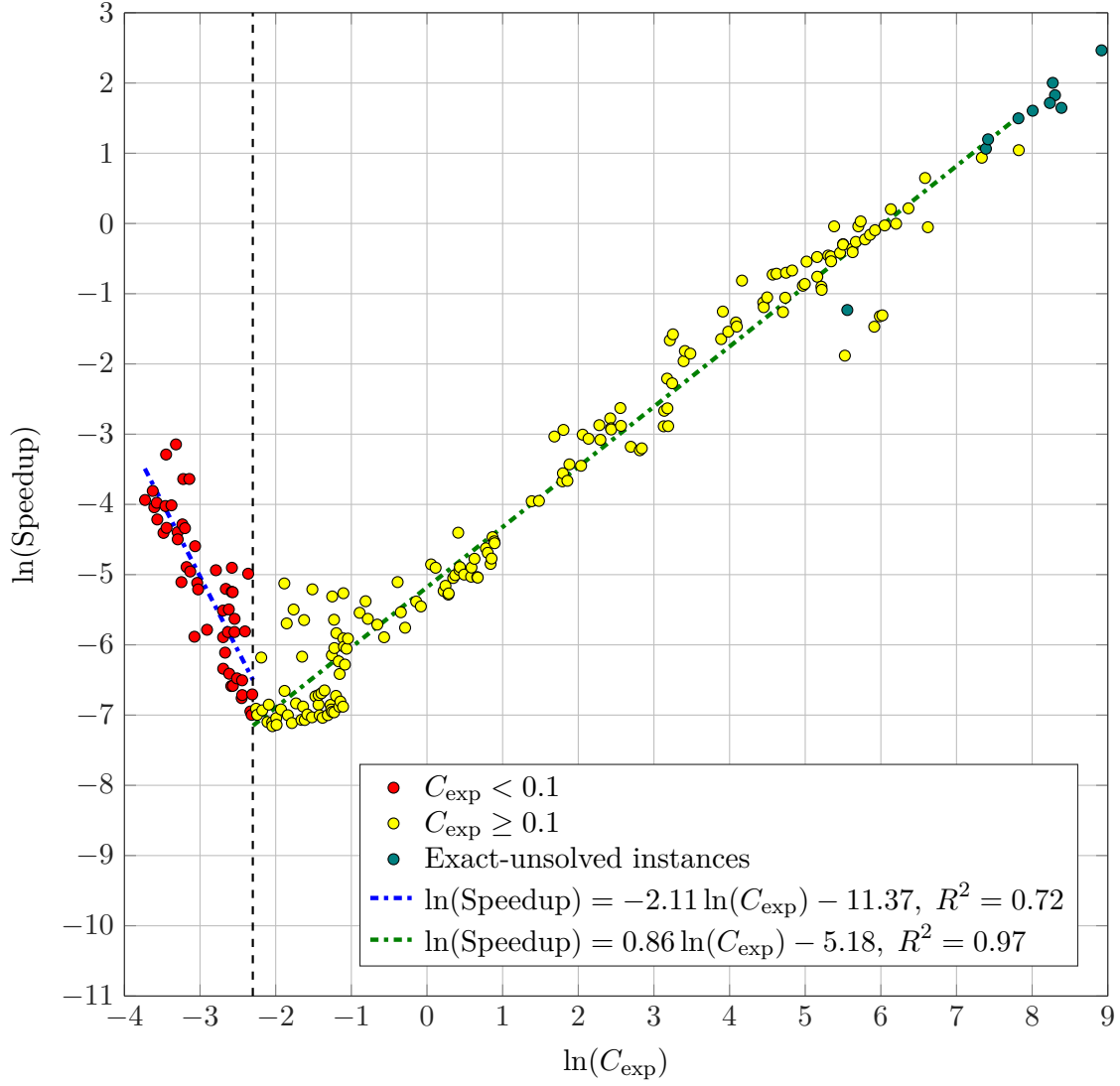
\begin{figure}[!htb]
\centering
\resizebox{1.00\linewidth}{!}{
\begin{tikzpicture}
\begin{axis}[%
grid=both,
width=5in,
height=5in,
xlabel={\(\ln(C_{\mathrm{exp}})\)},
ylabel={\(\ln(\mathrm{Speedup})\)},
scale only axis,
separate axis lines,
every outer x axis line/.append style={white!15!black},
every x tick label/.append style={font=\color{white!15!black}},
every outer y axis line/.append style={white!15!black},
every y tick label/.append style={font=\color{white!15!black}},
xmin=-4.0,
xmax=9,
ymin=-11.0,
ymax=3,
legend style={draw=white!15!black,legend cell align=left,legend pos=south east}
]

\addplot [
color=black,
only marks,
mark=*,
mark options={solid,fill=red},
restrict x to domain=-4.0:-2.302585,
unbounded coords=discard
]
table[
col sep=space,
x expr={ln(\thisrow{complexity})},
y expr={ln(\thisrow{speedup})}
]{complexity_speedup.dat};
\addlegendentry{\(C_{\mathrm{exp}}<0.1\)};

\addplot [
color=black,
only marks,
mark=*,
mark options={solid,fill=yellow},
restrict x to domain=-2.302585:8.3,
unbounded coords=discard
]
table[
col sep=space,
x expr={ln(\thisrow{complexity})},
y expr={ln(\thisrow{speedup})}
]{complexity_speedup.dat};
\addlegendentry{\(C_{\mathrm{exp}}\geq 0.1\)};

\addplot [
color=black,
only marks,
mark=*,
mark options={solid,fill=teal},
unbounded coords=discard
]
table[
row sep=crcr,
x expr={ln(\thisrow{complexity})},
y expr={ln(\thisrow{speedup})}
]{%
complexity speedup \\
258.914273626999 0.291691633423437 \\
1617.650520338 2.89974689805721 \\
1664.314399699 3.3149471455868 \\
2486.10260879199 4.4700182859678 \\
3001.09402998199 4.98133537278714 \\
4389.418053491 5.18781089948233 \\
4031.662073529 6.21164281173852 \\
3766.725155555 5.55966078681606 \\
3913.933334872 7.40809932491309 \\
7459.650330069 11.7679984528606 \\
9696.6268515 11.620950326773 \\
};
\addlegendentry{Exact-unsolved instances};

\addplot [
color=blue,
ultra thick,
dashdotted
]
table[row sep=crcr]{%
-3.73  -3.49 \\
-2.30  -6.50 \\
};
\addlegendentry{\(\ln(\mathrm{Speedup})=-2.11\ln(C_{\mathrm{exp}})-11.37,\;R^2=0.72\)};

\addplot [
color=green!50!black,
ultra thick,
dashdotted
]
table[row sep=crcr]{%
-2.30  -7.15 \\
7.83   1.53 \\
};
\addlegendentry{\(\ln(\mathrm{Speedup})=0.86\ln(C_{\mathrm{exp}})-5.18,\;R^2=0.97\)};

\addplot [
color=black,
dashed,
thick
]
table[row sep=crcr]{%
-2.302585  -11 \\
-2.302585   3 \\
};

\end{axis}
\end{tikzpicture}%
}
\caption{Relationship between the experimental complexity \(C_{\mathrm{exp}}\) and the speedup of the genetic algorithm. Both axes are represented in natural logarithmic scale. The vertical dashed line separates very small instances with \(C_{\mathrm{exp}}<0.1\) from the remaining instances. The red and yellow points correspond to exactly solved instances, while the teal points correspond to exact-unsolved instances for which the exact Bellman resolution time is estimated from the empirical regression. The two fitted regressions show two distinct regimes: a decreasing trend for very small instances, where GA overhead dominates, and an increasing trend for larger instances, where the GA becomes progressively more advantageous.}
\label{fig:experimental-complexity-speedup-log}
\end{figure}

We now analyze the relationship between the empirical complexity of an instance and the speedup obtained by the genetic algorithm. Since the full benchmark set contains heterogeneous instances, the complexity of an instance cannot be described by a single structural parameter such as the horizon length, the number of demands, or the demand quantity alone. We therefore define an experimental complexity index based on the computational effort required by the exact DTMDP approach.
For each instance, the experimental complexity is measured as $C_{\mathrm{exp}}=t_{\mathrm{build}}^{\mathrm{DTMDP}}+t_{\mathrm{exact}}^{\mathrm{DTMDP}}$, where \(t_{\mathrm{build}}^{\mathrm{DTMDP}}\) is the DTMDP construction time and \(t_{\mathrm{exact}}^{\mathrm{DTMDP}}\) is the exact finite-horizon resolution time. This quantity provides an empirical indicator of the actual computational difficulty of an instance for the exact method. The speedup is then analyzed as a function of this experimental complexity.
Figure~\ref{fig:experimental-complexity-speedup-log} reports the relationship between \(\ln(C_{\mathrm{exp}})\) and \(\ln(\mathrm{Speedup})\). The point cloud is divided into two regions. The first region corresponds to very small instances with \(C_{\mathrm{exp}}<0.1\), while the second region contains the remaining instances with \(C_{\mathrm{exp}}\geq 0.1\).

The results show two clearly distinct regimes. For very small instances, namely when \(C_{\mathrm{exp}}<0.1\), the regression line is decreasing. In this region, the exact DTMDP method is already extremely fast, while the GA still requires a fixed computational effort for initialization, fitness evaluation, crossover, mutation, and selection. As a result, small variations in the exact computational time can strongly affect the measured speedup, and the GA overhead dominates the comparison.
For instances with \(C_{\mathrm{exp}}\geq 0.1\), the behaviour changes. The regression becomes strongly increasing, with a high coefficient of determination. This indicates that, once the instance is sufficiently complex, the speedup grows almost regularly with the experimental complexity. In other words, the more expensive the exact DTMDP construction and resolution become, the more advantageous the GA becomes in relative terms. The teal points extend this interpretation to exact-unsolved instances. Their estimated speedups range from below one for the least favorable case to more than 11 for the largest predicted cases. This confirms that extrapolation should not be interpreted as a uniform guarantee: when the predicted exact solve time is still moderate, the GA overhead may remain significant, whereas the gain becomes substantial for the largest instances beyond exact solvability on the test machine.
This result is consistent with the previous experiments. The GA is not designed to outperform the exact method on very small instances, where dynamic programming can solve the problem almost immediately. Its advantage appears when the exact method starts to suffer from the growth of the state space, feasible action sets, and transition matrices. The experimental complexity index therefore provides a useful aggregate indicator: across heterogeneous instances, it captures the point at which the GA changes from being dominated by fixed algorithmic overhead to becoming computationally beneficial.

\subsection{Generation-level gap--speedup evolution on a difficult instance}
\label{sec:ga-generation-gap-speedup}

\begin{figure}[!htb]
\centering
\begin{subfigure}{0.48\linewidth}
\centering
\resizebox{\linewidth}{!}{
\begin{tikzpicture}
\begin{axis}[%
grid=both,
width=3.2in,
height=2.6in,
xlabel={Generation},
ylabel={Global gap (\%)},
xmin=0,
xmax=60,
ymin=0,
ymax=30,
scale only axis,
separate axis lines,
every outer x axis line/.append style={white!15!black},
every x tick label/.append style={font=\color{white!15!black}},
every outer y axis line/.append style={white!15!black},
every y tick label/.append style={font=\color{white!15!black}}
]
\addplot [color=black,mark=*,mark options={solid,fill=red}]
table[row sep=crcr]{%
generation gap \\
1 27.6055025271195 \\
2 27.6055025271195 \\
3 12.5606103883231 \\
4 12.5606103883231 \\
5 12.5606103883231 \\
6 12.5606103883231 \\
7 11.2716304211369 \\
8 9.85463758605679 \\
9 7.73862431615966 \\
10 7.73862431615966 \\
11 7.73862431615966 \\
12 7.73862431615966 \\
13 7.73862431615966 \\
14 7.73862431615966 \\
15 7.73862431615966 \\
16 7.73862431615966 \\
17 7.73862431615966 \\
18 7.73862431615966 \\
19 7.73862431615966 \\
20 7.73862431615966 \\
21 7.73862431615966 \\
22 7.73862431615966 \\
23 7.73862431615966 \\
24 7.73862431615966 \\
25 7.73862431615966 \\
26 7.73862431615966 \\
27 7.73862431615966 \\
28 7.73862431615966 \\
29 7.73862431615966 \\
30 7.73862431615966 \\
31 7.73862431615966 \\
32 7.73862431615966 \\
33 7.73862431615966 \\
34 7.73862431615966 \\
35 7.73862431615966 \\
36 7.73862431615966 \\
37 7.73862431615966 \\
38 7.73862431615966 \\
39 7.73862431615966 \\
40 7.73862431615966 \\
41 7.73862431615966 \\
42 7.73862431615966 \\
43 7.73862431615966 \\
44 7.73862431615966 \\
45 6.36970861321564 \\
46 5.42340378539684 \\
47 5.42340378539684 \\
48 5.42340378539684 \\
49 5.42340378539684 \\
50 5.42340378539684 \\
51 5.42340378539684 \\
52 5.42340378539684 \\
53 5.42340378539684 \\
54 4.34253968403784 \\
55 4.34253968403784 \\
56 4.34253968403784 \\
57 4.34253968403784 \\
58 4.34253968403784 \\
59 4.34253968403784 \\
60 4.34253968403784 \\
};
\end{axis}
\end{tikzpicture}%
}
\caption{Global optimality gap}
\end{subfigure}
\hfill
\begin{subfigure}{0.48\linewidth}
\centering
\resizebox{\linewidth}{!}{
\begin{tikzpicture}
\begin{axis}[%
grid=both,
width=3.2in,
height=2.6in,
xlabel={Generation},
ylabel={Cumulative speedup},
xmin=0,
xmax=60,
ymin=0,
ymax=60,
scale only axis,
separate axis lines,
every outer x axis line/.append style={white!15!black},
every x tick label/.append style={font=\color{white!15!black}},
every outer y axis line/.append style={white!15!black},
every y tick label/.append style={font=\color{white!15!black}}
]
\addplot [color=black,only marks,mark=*,mark options={solid,fill=yellow}]
table[row sep=crcr]{%
generation speedup \\
1 56.2661677616599 \\
2 48.5306001191649 \\
3 42.7780886003623 \\
4 38.4327621013774 \\
5 34.5628743626861 \\
6 31.5717702476232 \\
7 29.1630343654819 \\
8 27.0507590455351 \\
9 25.2484668632281 \\
10 23.5351398880039 \\
11 22.1454241777897 \\
12 20.8735792505836 \\
13 19.7507890886919 \\
14 18.751938857128 \\
15 17.7246343744795 \\
16 16.8256822393888 \\
17 15.9101418021547 \\
18 15.0283887300475 \\
19 14.1404891260445 \\
20 13.3979481530052 \\
21 12.7267594794975 \\
22 12.0902154439237 \\
23 11.5119056692958 \\
24 11.0005331334186 \\
25 10.5588158108946 \\
26 10.126832411381 \\
27 9.73738030617004 \\
28 9.37553568088099 \\
29 9.0281442831401 \\
30 8.71555636947886 \\
31 8.42476961479461 \\
32 8.15438296233261 \\
33 7.8801864253715 \\
34 7.637103909989 \\
35 7.41502429217652 \\
36 7.19303293804059 \\
37 6.98505561306877 \\
38 6.79666750061293 \\
39 6.61691993619267 \\
40 6.44195617058387 \\
41 6.28242298148561 \\
42 6.13124113644617 \\
43 5.9797603985152 \\
44 5.84199226123411 \\
45 5.71078591445241 \\
46 5.62580168492713 \\
47 5.53665723820773 \\
48 5.45709583282668 \\
49 5.37951563503581 \\
50 5.30359887604257 \\
51 5.22024769175317 \\
52 5.14369326211747 \\
53 5.06262932697317 \\
54 4.97586269782129 \\
55 4.88262665765017 \\
56 4.78791002471241 \\
57 4.69759173285171 \\
58 4.61196102037993 \\
59 4.52287236221159 \\
60 4.44255488993111 \\
};
\end{axis}
\end{tikzpicture}%
}
\caption{Cumulative exact/GA speedup}
\end{subfigure}
\caption{Generation-level evolution of the global gap and cumulative speedup for on difficult instance. The speedup compares the exact Bellman resolution time with the cumulative GA time up to each generation.}
\label{fig:ga-generation-gap-speedup}
\end{figure}
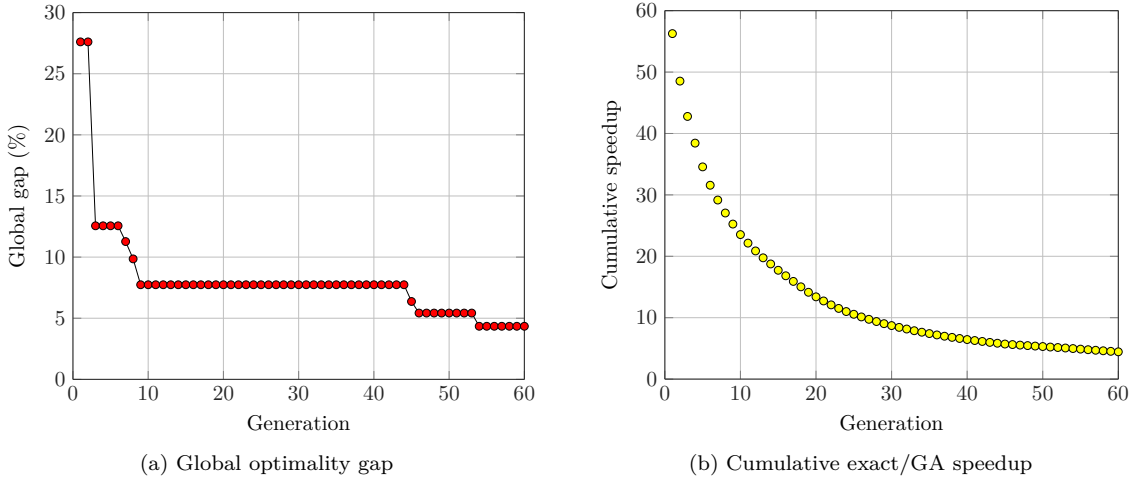

To illustrate how the GA progresses during a run, we analyze one difficult instance. This instance has horizon \(T=90\), one demand of quantity \(Q=100\), \(9923\) DTMDP states, and \(101\) actions. The exact DTMDP resolution succeeds on the test machine and gives an optimal expected cost of \(167.43\) in \(272.63\) seconds, which makes it suitable for observing the generation-level trade-off between solution quality and computation time.
Figure~\ref{fig:ga-generation-gap-speedup} reports the global optimality gap and the cumulative speedup at each generation. The gap decreases sharply during the first generations, from approximately \(27.61\%\) to \(7.74\%\) by generation 9. It then remains on a plateau for several generations before dropping to \(6.37\%\) at generation 45 and to \(5.42\%\) at generation 46. In parallel, the cumulative speedup necessarily decreases as more GA generations are evaluated: it starts above \(56\) after the first generation and remains above \(4.4\) at generation 60. This example highlights the practical stopping trade-off: early generations provide very large speedups but coarse policies, while later generations improve the gap at the cost of additional computation time.

\subsection{Sensitivity analysis of the holding--backlog cost ratio}
\label{sec:sensitivity-holding-backlog}

We perform a sensitivity experiment to analyze how the production policy reacts to the relative magnitude of holding and backlog costs. The objective is not only to compare costs, but also to study the temporal structure of the policy, namely whether production tends to be anticipated or delayed when the cost parameters change.
We consider a single-demand instance over a monthly horizon with daily periods. The planning horizon is fixed to \(T=30\). The demand quantity is \(Q_1=10\), and the demand can arrive within the window \([s_1,u_1]=[3,27]\). Its arrival distribution is generated as a smoothed unimodal distribution centered on day \(15\), using one peak at \(t=15\) and a smoothing parameter \(\sigma=3\). Thus, day \(15\) is the most likely arrival date, while the demand may still occur earlier or later within the admissible window.
All production parameters are kept fixed during the experiment. We use a daily capacity equal to \(20\), a setup time equal to \(1\), a unit production time equal to \(1\), a setup cost equal to \(1\), and a unit production cost equal to \(1\). The backlog cost is fixed to \(b=1\), while the holding cost is varied through the ratio
$\rho=\frac{h}{b}.$
Changing \(\rho\) therefore changes the relative cost of producing before the demand arrival compared with leaving the demand unmet after its arrival.
 For each value of \(\rho\), the genetic algorithm is used to compute a policy, with a population size of \(80\), \(150\) generations, the gene-uniform crossover, intelligent initialization, a mutation rate equal to \(0.05\), and a mutation gene fraction equal to \(0.1\).
To summarize the temporal behavior of a policy, we compute the expected production time. Let \(Z_t^\pi\) denote the production quantity decided at period \(t\) under policy \(\pi\). The expected production time is defined as
\begin{equation}
\bar{t}^{\pi}_{\mathrm{prod}}
=
\frac{\sum_{t=1}^{T} t\,\mathbb{E}[Z_t^\pi]}
{\sum_{t=1}^{T} \mathbb{E}[Z_t^\pi]}.
\end{equation}
This indicator can be interpreted as the average period at which units are produced under the stochastic evolution of the system and the policy returned by the genetic algorithm.

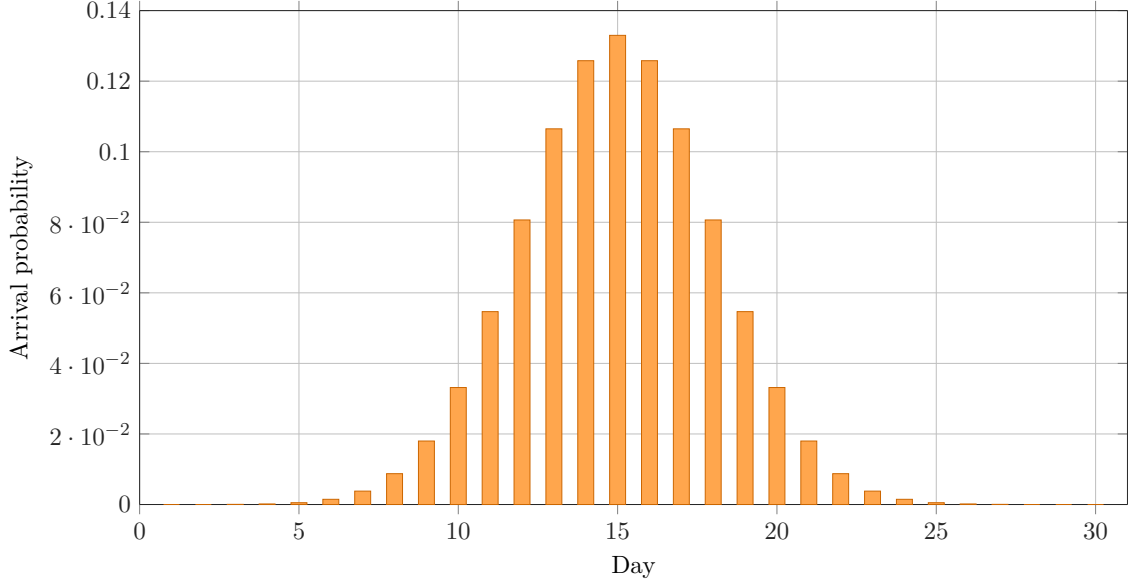
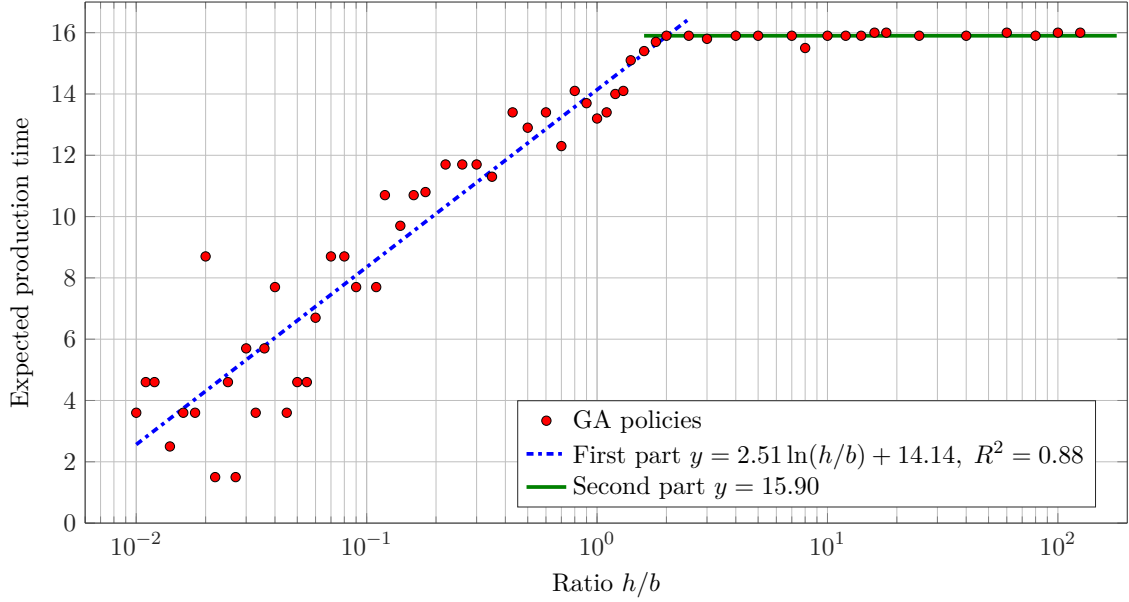
\begin{figure}[p]
\centering

\begin{subfigure}{1\linewidth}
\centering
\resizebox{1\linewidth}{!}{
\begin{tikzpicture}
\begin{axis}[
ybar,
bar width=7pt,
width=6in,
height=3in,
grid=both,
xlabel={Day},
ylabel={Arrival probability},
xmin=0,
xmax=31,
ymin=0,
ymax=0.14,
xtick={0,5,10,15,20,25,30},
ytick={0,0.02,0.04,0.06,0.08,0.10,0.12,0.14},
scale only axis,
separate axis lines,
every outer x axis line/.append style={white!15!black},
every outer y axis line/.append style={white!15!black},
every x tick label/.append style={font=\color{white!15!black}},
every y tick label/.append style={font=\color{white!15!black}},
]
\addplot[
fill=orange!70,
draw=orange!80!black
]
table[row sep=crcr]{
1   0.000000 \\
2   0.000000 \\
3   0.000045 \\
4   0.000160 \\
5   0.000514 \\
6   0.001477 \\
7   0.003799 \\
8   0.008741 \\
9   0.017998 \\
10  0.033160 \\
11  0.054672 \\
12  0.080659 \\
13  0.106486 \\
14  0.125798 \\
15  0.132985 \\
16  0.125798 \\
17  0.106486 \\
18  0.080659 \\
19  0.054672 \\
20  0.033160 \\
21  0.017998 \\
22  0.008741 \\
23  0.003799 \\
24  0.001477 \\
25  0.000514 \\
26  0.000160 \\
27  0.000045 \\
28  0.000000 \\
29  0.000000 \\
30  0.000000 \\
};
\end{axis}
\end{tikzpicture}
}
\caption{Arrival probability distribution of the single demand. The demand can arrive within the window \([3,27]\), and the distribution is unimodal and centered around day \(15\).}
\label{fig:single-demand-arrival-distribution}
\end{subfigure}

\vspace{0.6cm}

\begin{subfigure}{1\linewidth}
\centering
\resizebox{1\linewidth}{!}{
\begin{tikzpicture}
\begin{axis}[%
grid=both,
width=6.2in,
height=3.1in,
xlabel={Ratio $h/b$},
ylabel={Expected production time},
xmode=log,
xmin=0.006,
xmax=200,
ymin=0,
ymax=17,
scale only axis,
separate axis lines,
every outer x axis line/.append style={white!15!black},
every x tick label/.append style={font=\color{white!15!black}},
every outer y axis line/.append style={white!15!black},
every y tick label/.append style={font=\color{white!15!black}},
legend style={draw=white!15!black,legend cell align=left,legend pos=south east}
]

\addplot [color=black,only marks,mark=*, mark options={fill=red}]
table[row sep=crcr]{%
0.010 3.6 \\
0.011 4.6 \\
0.012 4.6 \\
0.014 2.5 \\
0.016 3.6 \\
0.018 3.6 \\
0.020 8.7 \\
0.022 1.5 \\
0.025 4.6 \\
0.027 1.5 \\
0.030 5.7 \\
0.033 3.6 \\
0.036 5.7 \\
0.040 7.7 \\
0.045 3.6 \\
0.050 4.6 \\
0.055 4.6 \\
0.060 6.7 \\
0.070 8.7 \\
0.080 8.7 \\
0.090 7.7 \\
0.110 7.7 \\
0.120 10.7 \\
0.140 9.7 \\
0.160 10.7 \\
0.180 10.8 \\
0.220 11.7 \\
0.260 11.7 \\
0.300 11.7 \\
0.350 11.3 \\
0.430 13.4 \\
0.500 12.9 \\
0.600 13.4 \\
0.700 12.3 \\
0.800 14.1 \\
0.900 13.7 \\
1.000 13.2 \\
1.100 13.4 \\
1.200 14.0 \\
1.300 14.1 \\
1.400 15.1 \\
1.600 15.4 \\
1.800 15.7 \\
2.000 15.9 \\
2.500 15.9 \\
3.000 15.8 \\
4.000 15.9 \\
5.000 15.9 \\
7.000 15.9 \\
8.000 15.5 \\
10.000 15.9 \\
12.000 15.9 \\
14.000 15.9 \\
16.000 16.0 \\
18.000 16.0 \\
25.000 15.9 \\
40.000 15.9 \\
60.000 16.0 \\
80.000 15.9 \\
100.000 16.0 \\
125.000 16.0 \\
};
\addlegendentry{GA policies};

\addplot [color=blue,ultra thick,dashdotted]
table[row sep=crcr]{%
0.01    2.56260215242594  \\
0.011   2.80221194445401  \\
0.012   3.02095854620594  \\
0.014   3.40849335529167  \\
0.016   3.74419127634972  \\
0.018   4.04029782798987  \\
0.02    4.30517416435364  \\
0.022   4.54478395638171  \\
0.025   4.86615705235757  \\
0.027   5.05963710977379  \\
0.03    5.32451344613757  \\
0.033   5.56412323816564  \\
0.036   5.78286983991757  \\
0.04    6.04774617628135  \\
0.045   6.34385272792149  \\
0.05    6.60872906428527  \\
0.055   6.84833885631334  \\
0.06    7.06708545806527  \\
0.07    7.454620267151  \\
0.08    7.79031818820905  \\
0.09    8.08642473984919  \\
0.11    8.59091086824104  \\
0.12    8.80965746999297  \\
0.14    9.1971922790787 \\
0.16    9.53289020013675  \\
0.18    9.8289967517769 \\
0.22    10.3334828801687  \\
0.26    10.7534568490119  \\
0.3   11.1132123699246  \\
0.35    11.5007471790103  \\
0.43    12.0182592432796  \\
0.5   12.3974279880723  \\
0.6   12.8557843818523  \\
0.7   13.243319190938 \\
0.8   13.5790171119961  \\
0.9   13.8751236636362  \\
1   14.14 \\
1.1   14.3796097920281  \\
1.2   14.59835639378  \\
1.3   14.7995837608713  \\
1.4   14.9858912028657  \\
1.6   15.3215891239238  \\
1.8   15.6176956755639  \\
2   15.8825720119277  \\
2.5   16.4435548999316  \\
};
\addlegendentry{First part $y=2.51\ln(h/b)+14.14,\;R^2=0.88$};

\addplot [
color=green!50!black,
ultra thick
]
table[row sep=crcr]{%
1.6 15.9  \\
2.5 15.9  \\
3 15.9  \\
4 15.9  \\
5 15.9  \\
7 15.9  \\
8 15.9  \\
10  15.9  \\
12  15.9  \\
14  15.9  \\
16  15.9  \\
18  15.9  \\
25  15.9  \\
40  15.9  \\
60  15.9  \\
80  15.9  \\
100 15.9  \\
180 15.9  \\
};
\addlegendentry{Second part $y=15.90$};

\end{axis}
\end{tikzpicture}
}
\caption{Expected production time obtained by the GA policies as a function of the holding-to-backlog cost ratio \(h/b\). The scatter points correspond to GA policies, while the two fitted segments highlight two regimes: an increasing regime for low values of \(h/b\), followed by a nearly saturated regime in which the expected production time becomes almost constant.}
\label{fig:holding-backlog-joint}
\end{subfigure}

\caption{Demand-timing experiment. The first panel reports the arrival probability distribution of the single demand, while the second panel shows how the expected production time obtained by the GA policies changes with the holding-to-backlog cost ratio.}
\label{fig:demand-distribution-and-production-time}
\end{figure}

Figure~\ref{fig:demand-distribution-and-production-time} reports the results of the experiment. The first panel shows the demand arrival distribution, which is concentrated around the middle of the horizon. The second panel shows the expected production time obtained for different values of \(\rho\). The results exhibit a clear increasing trend. When \(\rho\) is small, holding inventory is inexpensive relative to backlog penalties. The policy therefore tends to produce earlier, sometimes well before the most likely arrival date, in order to hedge against uncertainty in the demand arrival time. Pre-producing inventory reduces the risk of costly backlog periods if the demand materializes earlier than expected.

As \(\rho\) increases, holding inventory becomes more expensive. The policy progressively delays production, and the expected production time moves toward the center of the arrival distribution. Around intermediate values of \(\rho\), the transition is particularly visible: the expected production time shifts from early production periods toward days close to the most likely arrival date. This confirms that the model reacts coherently to the cost structure.

For large values of \(\rho\), the curve reaches a plateau around day \(16\). This saturation effect is expected. Once holding becomes sufficiently expensive, the policy no longer has an incentive to produce far in advance. However, because the demand is most likely to arrive around day \(15\), and because it must be satisfied no later than \(u_1=27\), the policy stabilizes around a production timing close to the peak of the arrival distribution rather than delaying production indefinitely.


\section{Discussion}
\label{sec:discussion}

The computational results support three main observations. First, the DTMDP formulation provides a precise way to model stochastic demand timing, but this precision has a significant computational cost. The stochastic model requires explicit arrival-status variables and stochastic transition branches. The deterministic-counterpart experiment confirms this effect: when each arrival distribution is replaced by a deterministic arrival at its most likely period, the resulting exact dynamic program is much smaller and much faster. The average stochastic/deterministic total-time ratio is above 100 on the comparable instances.
Second, the exact DTMDP remains the appropriate method for small instances because it provides an optimality certificate and can solve these instances quickly. However, as the horizon, demand quantities, number of demands, then the transition structure grow and the exact method becomes limited by both time and memory. 
Third, the genetic algorithm provides a practical compromise. It does not provide an optimality certificate, but it evaluates policies exactly under the stochastic transition model and keeps the average gap close to the target accuracy level. On the difficult instances satisfying the 5\% gap limit, the average speedup is \(6.89\pm1.41\) at the 95\% confidence level, showing that the computational gain remains meaningful under a controlled quality requirement. Its advantage becomes clearer on instances where the exact optimization phase becomes expensive. For exact-unsolved instances, the empirical Bellman-time regression provides an estimated exact baseline and indicates that the GA can remain computationally attractive beyond the exact solvability limit of the test machine. The GA is therefore best interpreted as a scalable approximate policy-search method for stochastic-timing lot-sizing instances that are too large for routine exact resolution.
The main limitation of the present study is that the GA still relies on the constructed DTMDP for policy evaluation. Therefore, it reduces the optimization burden but does not completely remove the memory cost of constructing the stochastic model. Future work should investigate simulation-based or decomposed policy-evaluation strategies that avoid storing the full transition model, as well as hybrid methods combining deterministic dynamic programming, stochastic approximation, and policy search.

\section{Conclusion}
\label{sec:conclusion}

We presented a discrete-time Markov decision process formulation for multi-item capacitated lot sizing with stochastic demand timing. The formulation models demands as individual objects with deterministic quantities and stochastic arrival periods. Demand-level production decisions allow the model to represent simultaneous arrivals, competition for limited capacity, demand-specific backlog, and demand-dedicated pre-production.
The deterministic-counterpart experiment highlights the specific cost of stochastic timing. Replacing each arrival distribution by a deterministic arrival at its most likely period leads to a much smaller deterministic dynamic program and substantially shorter exact solution times. This confirms that the stochastic timing component is a major driver of complexity in the proposed lot-sizing model.
The exact DTMDP formulation provides an optimal policy when the state-action-transition model can be constructed and solved. However, the computational study shows that stochastic timing can rapidly increase both the size of the transition model and the memory required for exact resolution. After establishing this deterministic-to-stochastic gap, we proposed a genetic algorithm that searches over feasible state-feedback policies and evaluates each policy exactly under the DTMDP transition model. On the tested benchmark instances, the GA remains close to the exact solution on average and becomes increasingly useful as instance size and exact computational effort increase. In particular, on the difficult instances we still satisfying the 5\% optimality-gap limit, and the GA reaches an average speedup of \(6.89\pm1.41\) at the 95\% confidence level. For instances that cannot be solved exactly on the available hardware, the empirical exact-time regression further provides a way to estimate the missing exact baseline and to extrapolate the likely speedup of the GA.
Future work will focus on reducing the dependence of the GA on the fully constructed DTMDP, extending the computational study to larger industrial instances, and designing decomposition or approximation methods that preserve the adaptive nature of the stochastic model while reducing memory requirements.

\bibliography{references}

@article{bayati2018power,
  author  = {Bayati, Marziyeh},
  title   = {Power Management Policy for Heterogeneous Data Center Based on Histogram and Discrete-Time MDP},
  journal = {Electronic Notes in Theoretical Computer Science},
  volume  = {337},
  pages   = {5--22},
  year    = {2018},
  doi     = {10.1016/j.entcs.2018.03.031}
}

@inproceedings{bayati2023discrete,
  author    = {Bayati, L{\'e}a},
  title     = {Discrete-Time MDP Policy for Energy-Aware Data Center},
  booktitle = {Proceedings of the 12th International Conference on Smart Cities and Green ICT Systems},
  pages     = {89--97},
  year      = {2023},
  doi       = {10.5220/0011846300003491}
}

@article{guner2010review,
  author  = {Guner Goren, Hacer and Tunali, Semra and Jans, Raf},
  title   = {A review of applications of genetic algorithms in lot sizing},
  journal = {Journal of Intelligent Manufacturing},
  volume  = {21},
  pages   = {575--590},
  year    = {2010},
  doi     = {10.1007/s10845-008-0205-2}
}

@article{xie2002heuristic,
  author  = {Xie, Jinxing and Dong, Jin},
  title   = {Heuristic genetic algorithms for general capacitated lot-sizing problems},
  journal = {Computers \& Mathematics with Applications},
  volume  = {44},
  pages   = {263--276},
  year    = {2002},
  doi     = {10.1016/S0898-1221(02)00146-3}
}

@article{toledo2013hybrid,
  author  = {Toledo, Claudio F. M. and de Oliveira, Renato R. R. and Fran{\c{c}}a, Paulo M.},
  title   = {A hybrid multi-population genetic algorithm applied to solve the multi-level capacitated lot sizing problem with backlogging},
  journal = {Computers \& Operations Research},
  volume  = {40},
  number  = {4},
  pages   = {910--919},
  year    = {2013},
  doi     = {10.1016/j.cor.2012.11.002}
}

@article{mohammadi2011genetic,
  author  = {Mohammadi, Mohammad and Fatemi Ghomi, Seyyed M. T.},
  title   = {Genetic algorithm-based heuristic for capacitated lot-sizing problem in flow shops with sequence-dependent setups},
  journal = {International Journal of Computer Integrated Manufacturing},
  year    = {2011},
  doi     = {10.1080/0951192X.2010.511654}
}

@article{babaei2014genetic,
  author  = {Babaei, Mohammad and Mohammadi, Mohammad and Ghomi, Seyed M. T. Fatemi},
  title   = {A genetic algorithm for the simultaneous lot sizing and scheduling problem in capacitated flow shop with complex setups and backlogging},
  journal = {The International Journal of Advanced Manufacturing Technology},
  volume  = {70},
  pages   = {125--134},
  year    = {2014},
  doi     = {10.1007/s00170-013-5252-y}
}

@article{wang2022adaptive,
  author  = {Wang, Shuai and Hui, Jizhuang and Zhu, Bin and Liu, Ying},
  title   = {Adaptive Genetic Algorithm Based on Fuzzy Reasoning for the Multilevel Capacitated Lot-Sizing Problem with Energy Consumption in Synchronizer Production},
  journal = {Sustainability},
  volume  = {14},
  number  = {9},
  pages   = {5072},
  year    = {2022},
  doi     = {10.3390/su14095072}
}

@article{Akartunali2022EJOR,
  author  = {Akartunal{\i}, K. and Dauz{\`e}re-P{\'e}r{\`e}s, S.},
  title   = {Dynamic lot sizing with stochastic demand timing},
  journal = {European Journal of Operational Research},
  year    = {2022},
  volume  = {302},
  number  = {1},
  pages   = {221--229}
}

@book{Puterman1994,
  author    = {Puterman, Martin L.},
  title     = {Markov Decision Processes: Discrete Stochastic Dynamic Programming},
  publisher = {Wiley},
  year      = {1994}
}

@article{wagner1958dynamic,
  title={Dynamic version of the economic lot size model},
  author={Wagner, Harvey M and Whitin, Thomson M},
  journal={Management science},
  volume={5},
  number={1},
  pages={89--96},
  year={1958},
  publisher={INFORMS}
}

@article{brahimi2017single,
  title={Single-item dynamic lot-sizing problems: An updated survey},
  author={Brahimi, Nadjib and Absi, Nabil and Dauz{\`e}re-P{\'e}r{\`e}s, St{\'e}phane and Nordli, Atle},
  journal={European Journal of Operational Research},
  volume={263},
  number={3},
  pages={838--863},
  year={2017},
  publisher={Elsevier}
}

@article{karimi2003capacitated,
  title={The capacitated lot sizing problem: a review of models and algorithms},
  author={Karimi, Behrooz and Ghomi, SMT Fatemi and Wilson, JM},
  journal={Omega},
  volume={31},
  number={5},
  pages={365--378},
  year={2003},
  publisher={Elsevier}
}

@article{quadt2008capacitated,
  title={Capacitated lot-sizing with extensions: a review},
  author={Quadt, Daniel and Kuhn, Heinrich},
  journal={4OR},
  volume={6},
  number={1},
  pages={61--83},
  year={2008},
  publisher={Springer}
}

@incollection{tempelmeier2013stochastic,
  title={Stochastic lot sizing problems},
  author={Tempelmeier, Horst},
  booktitle={Handbook of stochastic models and analysis of manufacturing system operations},
  pages={313--344},
  year={2013},
  publisher={Springer}
}

@article{aloulou2014bibliography,
  title={A bibliography of non-deterministic lot-sizing models},
  author={Aloulou, Mohamed Ali and Dolgui, Alexandre and Kovalyov, Mikhail Y},
  journal={International Journal of Production Research},
  volume={52},
  number={8},
  pages={2293--2310},
  year={2014},
  publisher={Taylor \& Francis}
}

@inproceedings{rodoplu2022multi,
  title={Multi Item Capacitated Lot Sizing with Stochastic Demand Timing},
  author={Rodoplu, Melek and Dauz{\`e}re-P{\'e}r{\`e}s, St{\'e}phane and Akartunali, Kerem},
  booktitle={23{\`e}me congr{\`e}s annuel de la Soci{\'e}t{\'e} Fran{\c{c}}aise de Recherche Op{\'e}rationnelle et d'Aide {\`a} la D{\'e}cision},
  year={2022}
}

@article{bibak2025integration,
  title={Integration of machine learning and optimization models for a data-driven lot sizing problem with random yield},
  author={Bibak, Bijan and Karaesmen, Fikri},
  journal={International Journal of Production Economics},
  volume={282},
  pages={109529},
  year={2025},
  publisher={Elsevier}
}

@inproceedings{metzker2021optimization,
  title={Optimization for Lot-Sizing Problems Under Uncertainty: A Data-Driven Perspective},
  author={Metzker, Paula and Thevenin, Simon and Adulyasak, Yossiri and Dolgui, Alexandre},
  booktitle={IFIP International Conference on Advances in Production Management Systems},
  pages={703--709},
  year={2021},
  organization={Springer}
}

@article{gong2023training,
  title={Training demand prediction models by decision error for two-stage lot-sizing problems},
  author={Gong, Hailei and Zhang, Yanzi and Zhang, Zhi-Hai},
  journal={IEEE Transactions on Automation Science and Engineering},
  volume={21},
  number={2},
  pages={1122--1137},
  year={2023},
  publisher={IEEE}
}

@article{deng2026computable,
  title={Computable Reformulation of Data-Driven Distributionally Robust Chance Constraints: Validated by Solution of Capacitated Lot-Sizing Problems},
  author={Deng, Hua and Wan, Zhong},
  journal={Mathematics},
  volume={14},
  number={2},
  pages={331},
  year={2026},
  publisher={MDPI}
}

\end{document}